%% file: SRS17-BA.tex
\documentclass[ba,preprint]{imsart2}
\usepackage{amsthm}
\usepackage{amsmath}
\usepackage{natbib}
\usepackage{graphics}
\usepackage{color}
\startlocaldefs
\numberwithin{equation}{section}
\theoremstyle{plain}

\endlocaldefs

\usepackage[dvipsnames]{xcolor}
\usepackage{amsfonts,amssymb}
\usepackage{enumerate}
\usepackage{natbib}
\usepackage{verbatim}
\usepackage{url} 
\usepackage{mathrsfs}
\usepackage{tikz}
\usetikzlibrary{arrows}
\usepackage{placeins}
\usepackage{float}
\usepackage{array}
\usepackage{stmaryrd}
\usepackage{braket}
\usepackage{enumitem}
\usepackage{caption}
\usepackage[caption=false]{subfig}
\usepackage{etex}
\usepackage{eufrak}

\def\R{{\mathbf R}}

\def\T{{\mathcal T}}
\def\F{{\mathcal F}}
\def\G{{\mathcal G}}
\def\E{{\mathcal E}}
\def\EE{{\mathbf E}}
\def\X{{\mathbf X}}
\def\x{{\mathbf x}}

\def\UU{{\mathcal U}}

\def\D{{\mathcal D}}
\def\V{{\mathbf V}}
\newcommand{\indep}{\mkern2mu \rotatebox[origin=c]{90}{$\models$}\mkern2mu}
\newcommand{\defeq}{\mathrel{\vcenter{\baselineskip0.5ex \lineskiplimit0pt
                     \hbox{\scriptsize.}\hbox{\scriptsize.}}}%
                     =}
\newtheorem{proposition}{Proposition}
\newtheorem{definition}{Definition}
\newtheorem{theorem}{Theorem}
\newtheorem{lemma}{Lemma}

\xdefinecolor{tree}{RGB}{162,192,217}
\xdefinecolor{ER2p}{RGB}{246,173,160}
\xdefinecolor{ER4p}{RGB}{230,167,156}
\xdefinecolor{ER8p}{RGB}{197,154,146}
\newcommand\crule[3][black]{\textcolor{#1}{\rule{#2}{#3}}}

\newcommand{\SRnew}[2]{\textcolor{black}{#2}}
\newcommand{\LS}[2]{#2}

\begin{document}

\begin{frontmatter}

\title{A closed-form approach to Bayesian inference in tree-structured graphical models}
\runtitle{Tree-structured graphical models: a closed-form approach}


\begin{aug}
\author{\fnms{Lo\"ic} \snm{Schwaller}\thanksref{addr1,m1}\ead[label=e1]{loic.schwaller@ens-lyon.org}},
\author{\fnms{St\'ephane} \snm{Robin}\thanksref{addr2,m1,m2}
}
\and
\author{\fnms{Michael} \snm{Stumpf}\thanksref{addr3,m2}
}

\runauthor{L. Schwaller et al.}

\address[addr1]{Mathematical Institute, Leiden University, P.O. Box 9512, 2300 RA Leiden, The Netherlands
    \printead{e1} 
}

\address[addr2]{UMR MIA-Paris, AgroParisTech, INRA, Universit\'{e} Paris-Saclay, 75005 Paris, France 
}

\address[addr3]{Centre for Integrative Systems Biology and Bioinformatics, Imperial College London, London, United Kingdom
}


\end{aug}

\begin{abstract}
\SRnew{ 
  We consider the inference of the structure of an undirected graphical model in an exact Bayesian framework. More specifically we aim at achieving the inference with close-form posteriors, avoiding any sampling step. This task would be intractable without any restriction on the considered graphs, so we limit our exploration to {mixtures of} spanning trees.
  }{
  We consider the inference of the structure of an undirected graphical model in a Bayesian framework. To avoid convergence issues and highly demanding Monte Carlo sampling, we focus on exact inference. More specifically we aim at achieving the inference with close-form posteriors, avoiding any sampling step. To this aim, we restrict the set of considered graphs to {mixtures of} spanning trees.
  } 
  We investigate under which conditions on the priors --~on both tree structures and parameters~-- exact Bayesian inference can be achieved. Under these conditions, we derive a fast an exact algorithm to compute the posterior probability for an edge to belong to {the tree model} using an algebraic result called the Matrix-Tree theorem. We show that the assumption we have made does not prevent our approach to perform well on synthetic and flow cytometry data.
\end{abstract}



\end{frontmatter}

\input{tex/intro.tex}

\input{tex/model.tex}

\input{tex/priors.tex}

\input{tex/inference.tex}

\input{tex/simuls.tex}

\input{tex/appli.tex}

\input{tex/appendix.tex}

\begin{acknowledgement}
 \SRnew{}{The authors thank Sophie {\sc Donnet} for her helpful comments and remarks.}
\end{acknowledgement}

\bibliographystyle{ba}
\bibliography{bibliography3}

\end{document}

%% file: tex/intro.tex
\section{Introduction}
Statistical models are getting more and more complex and can now involve very intricate dependency structures. Graphical models are both a natural and powerful way to depict such structures. Inferring a graphical model based on observed data is hence of great interest for many fields of applications. 
From a statistical point-of-view, considering the inference of a graphical model requires to consider the graphical model itself as a parameter. In a Bayesian context, it means that we have to define a full model and, more specifically, a prior distribution on graphical models, therefore on graphs themselves.

Regardless of whether we consider directed or undirected graphs, their sheer number make them difficult to deal with. Markov Chain Monte Carlo (MCMC) methods have for instance  been used to sample from some sets of graphs, such as Directed Acyclic Graphs (DAGs) \citep{Madigan1995,Friedman2003,NiinimakiPK11} or decomposable graphs \citep{Green2013}. The decomposability assumption for undirected graphical models, also called Markov random fields, is commonly made in the literature, although some interest has been devoted to the less easy to handle non-decomposable graphs \citep{Roverato2002,Atay-Kayis2005}. The sampling schemes developed in the aforementioned papers are often subject to standard issues related to MCMC sampling in high-dimensional spaces, namely slow mixing and difficulty to get to the stationary distribution. This motivates our choice to focus on closed-form inference whenever possible.

In this paper, we refer to exact Bayesian inference, as Bayesian inference that \SRnew{}{does not rely on a sample from the posterior distribution but} provides closed-form posterior distributions of the parameters of interest, without sampling step. 
\SRnew{
{Such} approaches do not apply for arbitrary graph structures due to the combinatorial complexity. If no restriction is made on the topology of the graphical model, {closed-form} inference can only be contemplated as long as there are no more than thirty or so variables of interest \citep{Parviainen2009}. 
{For larger problems, closed form} approaches can only be considered at the price of a restriction on the structure space. When a subset of graphs is considered, it sometimes becomes possible to get access to the full posterior distribution on graphs. The obvious drawback of this approach is that the ``true'' graph might not belong to this subset. In this case, computing a maximum a posteriori (MAP) estimate {of the whole graph} would for instance yield a systematically wrong answer. {However}, such methods are not intended to assess the global structure all at once but to separately assess a collection of local features of the graph (typically, edges). The idea is that the inference of such features is less affected by the restriction than the global structure. {Still, as shown hereafter, Bayesian inference for such local features requires to integrate over the whole set of considered graphs.} In that perspective, trees have been of particular interest as a subset of both decomposable graphs and DAGs \citep{Chow68,Meila2001,Meila06,Kirshner2008,Lin2009,Burger2010}.
}{
Theoretically, closed-form posterior distributions on graphs can be computed, but the combinatorial complexity becomes prohibitive as soon as there are more than thirty or so variables of interest \citep{Parviainen2009}.
{For larger problems, closed-form} approaches can be considered at the price of a restriction on the structure space. When a subset of graphs is considered, it becomes possible to get access to the full posterior distribution on graphs, provided that the integration over the whole space of graphs can be achieved with a reasonable computational burden. In that perspective, trees have been of particular interest as a subset of both decomposable graphs and DAGs \citep{Chow68,Meila2001,Meila06,Kirshner2008,Lin2009,Burger2010}.
}

In this paper, we consider tree-based structure inference and we discuss under which conditions exact Bayesian inference can be achieved.
Our first contribution is to provide a well-defined fully Bayesian framework for graphical model inference based on trees.  We use the work of \cite{Dawid1993} on hyper Markov laws to define priors on tree parameters and distributions that can easily be marginalised over. This framework spares us from requiring likelihood equivalence between Markov-equivalent directed tree models, like \cite{Meila06} did building on the work of \cite{Heckerman1993}. We also point out that it fits within the recent work of \cite{Byrne2015} on structurally Markov graph distributions. We then go through a series of typical models befitting this framework, namely tree-structured copulas \citep{Kirshner2008}, multinomial distributions \citep{Meila06} and Gaussian distributions. Bayesian inference in this framework requires integration over the set of trees, that can be carried out exactly and efficiently using an algebraic result called the Matrix-Tree theorem.

Our second contribution focuses on edge inference. When \cite{Meila06} and \cite{Kirshner2008} were interested in the joint distribution of the observations, we are interested in the inference of the dependence structure. To this purpose, we are not concerned with the inference of the parameters but we need to account for the uncertainty of their estimates. The Bayesian construction we propose provides a natural framework to achieve this. We derive the exact posterior probability of any given edge, allowing for an arbitrary prior edge appearance probability. \\
Most works on tree-structured graphical model inference rely on the aforementioned Matrix-Tree theorem. As noticed by \cite{Kirshner2008}, the computation of posterior probabilities for all the edges in this setting can be achieved with cubic complexity with respect to the number of variables. We provide a new proof of this result relying on a generalization of the Matrix-Tree theorem to forests. 
{This enables us to derive a series of new results about the exact calculation of posterior characteristics such as the entropy of the posterior distribution of the tree or the posterior mean and variance of the degree of each node.}

\SRnew{
Our last contribution is a simulation study which addresses the influence of the tree assumption on the accuracy of structure inference for non-tree-structured graphical models. We demonstrate that, as long as edge inference is concerned, the computational efficiency following from this assumption can be obtained at a limited cost.
}{
Our last contribution is a simulation study which addresses the influence of the tree assumption on the accuracy of structure inference for non-tree-structured graphical models. 
Indeed, the `true' graph is unlikely to be a spanning tree, so computing a maximum a posteriori (MAP) estimate {of the whole graph} would for instance yield a systematically wrong answer. {However}, our approach is not designed to assess the global structure all at once but to separately assess a collection of local features of the graph (typically, edges). 
The rationale is that the inference of such features is weakly affected by the restriction to spanning tree. 
In the simulation study, we demonstrate that, as long as edge inference is concerned, the tree-based approach provides similar results as this obtained when considering a larger class of graphs, but with a dramatic reduction of the computational time.
}

An \textsf{R}-language package \textbf{saturnin} implementing the approach presented here is available  from the Comprehensive \textsf{R} Archive Network {at \url{https://cran.r-project.org/web/packages/saturnin/}}.

In Section \ref{sec_backgroung_model}, we provide some background on graphical models and Markov properties before writing down the full model in which the inference is performed. Priors for tree structures and distributions are defined in Section \ref{sec_apriori}. Section \ref{sec_inference} deals with the inference of the model. Integrations with respect to distributions and structures are respectively discussed in Sections \ref{subsec_pi_int} and \ref{subsec_matrixtree}. The simulation study and its results are described in Section \ref{sec_simu}. An application to flow cytometry data is presented in Section \ref{sec_app}.

%% file: tex/model.tex
\section{Background \& model}
\label{sec_backgroung_model}

\subsection{Markov properties \& graphical models}

Let $V = \{1,...,p\}$ and let $\X = (X_1,...,X_p)$ be a random vector indexed by $V$ and taking values in a product space $\mathcal{X} = \bigotimes_{i = 1}^p \mathcal{X}_i$. We let $\F$ denote the set of distributions on $\mathcal{X}$. For any subset $A$ of $V$, $\X_A$ stands for the subvector of $\X$ indexed by $A$. We also let $\mathcal{P}_2(V)$ denote the subsets of $V$ of size $2$. For $E\subseteq \mathcal{P}_2(V)$, $G = (V,E)$ is the undirected graph with vertices $V$ and edges $E$. In the following, the notations of \cite{Dawid1993} will be used. We refer the reader to the appendix of their article for a quick introduction to graph terminology and graphical models, or to \citep{Lauritzen1996} for a more detailed overview.\\
A pair $(A,B)$ of subsets of $V$ is said to be a decomposition of $G$ if $V= A \cup B$, if the subgraph induced by $G$ on $A\cap B$ is complete and if $A\cap B$ separates $A$ from $B$. When $A$ and $B$ are both proper subsets of $V$, the decomposition is said to be proper. Here we restrain our attention to decomposable graphs, namely graphs that are either complete or for which there exists a proper decomposition into two decomposable subgraphs.
 
\begin{definition}
A distribution $\pi \in \F$ is said to be Markov with respect to (w.r.t.) a decomposable graph $G$ if, for any decomposition $(A,B)$ of $G$, it holds that $\X_A \indep \X_B | \X_{A\cap B}$ under $\pi$.
\end{definition}

\begin{proposition}{\citep{HammersleyClifford71}}
Let $\pi \in \F$. If $\pi$ is a positive distribution (for all $\x \in \mathcal{X}$, $\pi(\x) > 0$), being Markov w.r.t. a decomposable graph $G$ is equivalent to the existence of a factorisation of $\pi$ on the (maximal) cliques of $G$.
\end{proposition}

We will focus on distributions that are Markov w.r.t. to connected graphs without any cycles. Such graphs are called spanning trees and their maximal cliques are of size $2$. Thus, a positive distribution that is Markov w.r.t. a tree $T=(V,E_T)$ can be factorised on the edges of the tree, using the marginal distributions of order $1$ and $2$:
  \begin{align*}
 \forall \x \in \mathcal{X},~ \pi(\x) &= \prod_{i \in V}\pi_i(x_i)\prod_{\{i,j\} \in E_T} \frac{\pi_{ij}(x_i,x_j)}{\pi_i(x_i)\pi_j(x_j)}.
 \end{align*}
Such distributions will be referred to as tree distributions in the following.

\begin{definition}
A graphical model  $m_G \defeq (G,\F_G)$ is given by a decomposable graph $G$ and a family of distributions $\F_G \subseteq \F$ that are Markov w.r.t. $G$. 
\end{definition}

Let $m_G = (G,\F_G)$ be a graphical model. To avoid any confusion, distributions on a set of distributions will be called hyperdistributions. For $\pi \in \F_G$ and $A,B \subseteq V$, we let $\pi_{A}$ denote the marginal distribution obtained from $\pi$ on the variables $\X_{A}$, and $\pi_{B|A}$ denote the collection of conditional distributions of $\X_{B} | \X_{A}$ under $\pi$. If $\rho$ is a hyperdistribution on $\F_G$, we also let $\rho_A$ and $\rho_{B|A}$ respectively denote the marginal hyperdistribution induced by $\rho$ on $\pi_A$ and the collection of hyperdistributions induced by $\rho$ on $\pi_{B|A}$. 
\begin{definition}
A hyperdistribution $\rho$ is said to be strong hyper Markov w.r.t. $G$ if, for any decomposition $(A,B)$ of $G$, it holds that $\pi_{A} \indep \pi_{B|A}$ under $\rho$.
\end{definition}Such hyperdistributions will be useful to define priors on distribution spaces.

\subsection{Model for Bayesian inference of graphical models based on trees}
\label{subsec:general_model}

Let $\T$ denote the set of spanning trees on $V$. For any tree $T\in \T$, we consider a graphical model $m_T = (T,\F_T)$ with a family of positive distributions $\F_T \subseteq \F$ Markov w.r.t. $T$. As we consider a Bayesian framework, we need to define prior distributions for $T$ and for $\pi$ conditionally on $T$. This is dealt with in Section \ref{sec_apriori}. The full Bayesian model is hierarchically described as follows. We first draw a random tree $T^*$ in the set of spanning trees, then a distribution $\pi$ in $\F_T$ and finally $\X$ according to $\pi$.
Defining a prior on tree distributions could be especially troublesome since it needs to be defined for every graphical model $m_T$. The idea is to require these hyperdistributions to be strong hyper Markov w.r.t. to their trees, so that they can be built from local hyperdistributions defined on the edges and chosen once and for all trees. This choice of priors and the fact that we only consider trees as possible structures make the inference of the graph in our model tractable in an exact manner.

%% file: tex/priors.tex
\section{Priors on tree structures \& distributions}
\label{sec_apriori}

The cardinality of $\T$ is $p^{p-2}$. Thus, restraining possible structures to spanning trees still leaves us with a large collection of graphical models to consider. Nonetheless, a suitable choice of priors on tree structures and parameters leads to a tractable situation. \cite{Meila06} define what they call decomposable priors under which parameters can be dealt with at the edge level. The integration over the set of trees can then be performed exactly using algebra. We will make use of strong hyper Markov hyperdistributions \citep{Dawid1993} to define our priors, but the idea is basically the same. Let $D = (\x^{(1)}, ... ,\x^{(n)})$ be an independent sample of size $n \geq 1$ drawn from $\X$. Our goal is to define a prior distribution $\xi$ on $(T,\pi)$ such that the posterior distribution on trees $\xi(\cdot|D)$ factorises over the edges, \textit{i.e.}
\begin{align}
\label{eq_fact_posterior}
\xi(T|D) &= \frac{1}{Z}\prod_{\{i,j\} \in E_T} \omega_{ij}, & \forall T\in \T,
\end{align}
where $\omega = (\omega_{ij})_{(i,j) \in V^2}$ is a symmetric matrix with non-negative values and $Z = \sum_{T\in \T} \prod_{\{i,j\} \in E_T} \omega_{ij}$ is a normalising constant. Both $\omega$ and $Z$ obviously depend on the data $D$, but we drop the dependence in the notations for the sake of clarity.

\subsection{Prior on tree structures}
\label{subsec:prior_structure}

Let $\beta = (\beta_{ij})_{(i,j) \in V^2}$ be a symmetric matrix with non-negative values such that the support graph $G_{\beta} = (V,E_{\beta})$ of $\beta$, where $E_{\beta} \defeq \left\lbrace \{i,j\} \in \mathcal{P}_2(V): \beta_{ij} > 0\right\rbrace$, is connected. We consider a prior distribution $ \xi$ on $\T$ that factorises over the edges,
\begin{align}
\label{prior_structure}
\xi(T) & = \frac{1}{Z_0} \prod_{\{i,j\} \in E_T} \beta_{ij}, & \forall T\in \T.
\end{align}
The assumption about $\beta$ is here to serve as a guarantee that $\beta$ induces a proper distribution on trees; $\xi$ can typically be taken as a uniform distribution on $\T$.

These distributions belong to the family of structurally Markov graph distributions described by \cite{Byrne2015} (see Section \ref{subsec:structurally_markov}).

\subsection{Prior on tree distributions}
\label{subsec_prior_param}

As Bayes' rule states that $\xi(T|D) \propto \xi(T)p(D|T)$, we are now interested in the marginal likelihood of the data under a tree model $m_T$,
\begin{align}
\label{marginal_likelihood}
p(D|T) = \int_{\F_T} p(D|\pi)p(\pi|T)d\pi.
\end{align}
For every $T \in \T$, we have to define a prior distribution on $\F_T$ such that the marginal likelihood $p(D|T)$ can also be factorised on the edges.

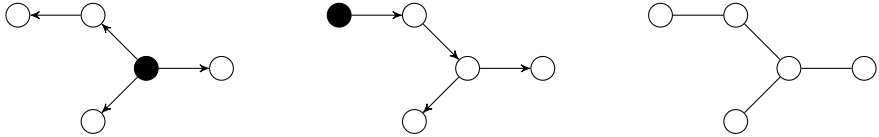
\begin{figure}[b]
 \centering
 
 \begin{tikzpicture}[->,>=stealth',auto,node distance=1cm,main node/.style={circle,draw,inner sep=0pt,minimum size=9pt}]
 
  		\node[main node,fill=black] (5) {};
 		\node[main node] (2) [above left of=5] {};
 		\node[main node] (1) [left of=2] {};
 		\node[main node] (3) [below left of=5] {};
 		\node[main node] (7) [right of=5] {};
 
 		\path[every node/.style={font=\sffamily\small}]
     		(2) edge node [left] {} (1)
     		(5) edge node [left] {} (2)
     		(5) edge node [left] {} (3)
     		(5) edge node [left] {} (7);
 		\end{tikzpicture}
 		\hspace{1cm}
 		\begin{tikzpicture}[->,>=stealth',auto,node distance=1cm,main node/.style={circle,draw,inner sep=0pt,minimum size=9pt}]
 
  		\node[main node] (5) {};
 		\node[main node] (2) [above left of=5] {};
 		\node[main node,fill=black] (1) [left of=2] {};
 		\node[main node] (3) [below left of=5] {};
 		\node[main node] (7) [right of=5] {};
 
 		\path[every node/.style={font=\sffamily\small}]
     		(1) edge node [left] {} (2)
     		(2) edge node [left] {} (5)
     		(5) edge node [left] {} (3)
     		(5) edge node [left] {} (7);
 		\end{tikzpicture}
 		\hspace{1cm}
 		\begin{tikzpicture}[-,>=stealth',auto,node distance=1cm,main node/.style={circle,draw,inner sep=0pt,minimum size=9pt}]
 
  		\node[main node] (5) {};
 		\node[main node] (2) [above left of=5] {};
 		\node[main node] (1) [left of=2] {};
 		\node[main node] (3) [below left of=5] {};
 		\node[main node] (7) [right of=5] {};
 
 		\path[every node/.style={font=\sffamily\small}]
     		(2) edge node [left] {} (1)
     		(5) edge node [left] {} (2)
     		(5) edge node [left] {} (3)
     		(5) edge node [left] {} (7);
 		\end{tikzpicture}
 		
 		\caption{Markov-equivalent directed and undirected trees.}
 		\label{fig:markov-eq}
 \end{figure}

\cite{Meila06} built their prior on multinomial tree distributions around three main assumptions, namely likelihood equivalence, parameter independence and parameter modularity. The first assumption requires that the prior treats all possible parametrisations consistent with a given tree $T$ (be it directed or undirected) as indistinguishable. These trees belong to the same equivalence class for the Markov equivalence relation. Moreover, this class only contains one undirected tree and several (namely $p$) directed trees. Figure \ref{fig:markov-eq} displays an example of Markov-equivalent trees.  Actually, all directed trees built from the undirected tree on the right panel by choosing a root and directing edges away from this root belong to the same Markov equivalence class. As a consequence, considering undirected graphs releases us from assuming likelihood equivalence. As for the parameter independence assumption, it can be broken down into local and global independences \citep{Spiegelhalter1990}. Strong hyper Markov hyperdistributions satisfy global independence but not necessarily local independence. The latter is in fact not needed to get the desired factorisation property for the marginal likelihood. Finally, the parameter modularity assumption is ensured by the construction of a compatible family of strong hyper Markov hyperdistributions.

\noindent Let $T$ be a tree and $\rho^T$ be a strong hyper Markov hyperdistribution on $\F_T$. Such hyperdistributions have an interesting property regarding the marginal likelihood $p(D|T)$.
\begin{proposition}
\label{markov_marginal}{\citep[Prop. 5.6]{Dawid1993}} If $\rho^T$ is strong hyper Markov w.r.t. $T$, then the marginal likelihood $p(D|T)$ is Markov w.r.t. to $T$.
\end{proposition}
This means that the marginal likelihood  can be factorised on the edges of $T$. For $i\in V$, let $D_i = \{x^{(1)}_i,...,x^{(n)}_i\}$ be the observed data restricted to $X_i$. The integral given in (\ref{marginal_likelihood}) can then be rewritten as
\begin{align}
p(D|T) = \int \pi(D)\rho^T(\pi)d\pi 
  = \prod_{i\in V}p(D_{i}|T)\prod_{\{i,j\} \in E_{T}}\frac{p(D_{i},D_{j}|T)}{p(D_{i}|T)p(D_{j}|T)}
 \label{eq:likelihood}
\end{align}
where, for all $(i,j)\in V^2$,
\begin{align}
 p(D_{i},D_{j}|T) &= \int \pi_{ij}(D_{i},D_{j})\rho^T_{ij}(\pi_{ij})d\pi_{ij}; \label{marginal_likelihood_2}\\ 
 p(D_{i}|T) &= \int \pi_i(D_{i})\rho^T_{i}(\pi_{i})d\pi_{i}. \nonumber
\end{align} 
The calculation of these integrals will be addressed in Section \ref{subsec_pi_int}.

We now explain how to choose $\rho^T$ for all $T$ so that the hyperdistributions of $\{\pi_{ij}\}_{\{i,j\}\in \mathcal{P}_2(V)}$ do not depend on $T$. Let us consider a general hyperdistribution $\rho$ on $\F$ such that, for any $A\subseteq V$, under $\rho$, 
\begin{align}
\pi_{A} \indep \pi_{V\setminus A | A}. \label{cond_comp}
\end{align}
This means that $\rho$ is strong hyper Markov w.r.t. the complete graph over $V$.
\begin{proposition}{\citep[\S 6.2]{Dawid1993}}
For any tree $T\in \T$, there exists a unique hyperdistribution $\rho^T$ on $\F_T$ that is strong hyper Markov w.r.t. $T$ and such that, for every edge $\{i,j\} \in E_T$, $\rho^T_{ij} = \rho_{ij}$. The collection $\{\rho^T\}_{T \in \T}$ is said to be a (hyper) compatible family of strong hyper Markov hyperdistributions.
\label{prop_compatible_family}
\end{proposition}

\begin{figure}[b!]
 \hfill\begin{minipage}{0.20\linewidth}\vfill
 \begin{align*}
 T^* &\sim \xi;\\
 \pi &\sim \rho^T;\\
 \X  &\sim \pi.
 \end{align*}
 \vfill
 \end{minipage}
 \hfill
 \begin{minipage}{0.64\linewidth}
 	\centering
 		\begin{tikzpicture}[->,>=stealth',shorten >=1pt,auto,node distance=1.3cm,
  		 thick,main node/.style={circle,fill=blue!0,draw,minimum size=24pt}];
  		 
 \tikzstyle{place}=[circle,draw=black!75,fill=black!10,minimum size=5mm]; 		 
  		 
 		\node[main node] (1)  {\footnotesize $\pi$};
 		\node[place] (2) [above of=1] {\footnotesize $\rho$};
 		\node[main node] (3) [left of=1] {\footnotesize $T^*$};
   		\node[main node] (4) [right of=1] {\footnotesize $\X$};
   		\node[place] (5) [above of=3] {\footnotesize $\beta$};

 		\path[every node/.style={font=\sffamily\small}]
     		(2) edge node {} (1)
     		(3) edge node {} (1)
     		(5) edge node {} (3)
 		(1) edge node {} (4);
 		\end{tikzpicture}\\
 \end{minipage}	
 	
 	\caption{Compatible strong hyper Markov tree model.}
 	\label{treemodel}
 \end{figure}
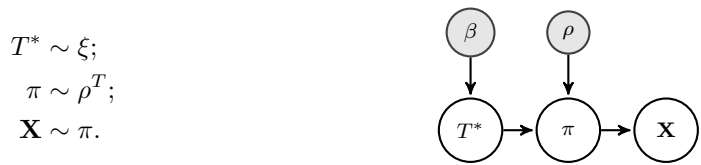

Proposition \ref{prop_compatible_family} guarantees that all $\rho^T$ are strong hyper Markov w.r.t. $T$. By Proposition \ref{markov_marginal}, for all $T\in \T$, the marginal likelihood under $\rho^T$ is Markov w.r.t. $T$. Moreover, the compatibility of the family $\{\rho^T\}_{T \in \T}$ makes the dependence on $T$ in the local marginal distributions given in (\ref{marginal_likelihood_2}) irrelevant. They can be computed once and for all for every $\{i,j\} \in \mathcal{P}_2(V)$. With this choice of hyperdistributions, the factorisation property needed for the posterior tree distribution (Eq. \ref{eq_fact_posterior}) is satisfied with
\begin{align}
\label{omega}
\omega_{ij} &= \beta_{ij}\frac{p(D_{i},D_{j})}{p(D_{i})p(D_{j})}, & \forall (i,j)\in V^2.
\end{align}
A full description of the model is given in Figure \ref{treemodel}.

Proposition \ref{prop_compatible_family} shows that we do not need to have access to the full basis hyperdistribution to specify a compatible family of strong hyper Markov hyperdistributions. It is indeed enough to provide a consistent family of pairwise hyperdistributions $\{\rho_{ij}\}_{\mathcal{P}_2(V)}$, where the consistency property must be understood in the sense that two hyperdistributions involving a common vertex should induce the same marginal hyperdistribution on this vertex. This is automatically satisfied when $\{\rho_{ij}\}_{\{i,j\} \in\mathcal{P}_2(V)}$ is obtained from a fully specified hyperdistribution $\rho$. In order to obtain strong hyper Markov hyperdistributions when combining these pairwise hyperdistributions, we shall additionally require that, for all $i,j \in V$, $\pi_{i|j} \indep \pi_{j}$ under $\rho_{ij}$ \citep[Prop. 3.16]{Dawid1993}, meaning that $\rho_{ij}$ is strong hyper Markov w.r.t. the graph on $\{i,j\}$ where vertices $i$ and $j$ are connected.

\subsection{Structural Markov property and structurally meta Markov families}
\label{subsec:structurally_markov}

The purpose of this section is to show how the model that we have described so far is related to the structural Markov property defined by \cite{Byrne2015}. Indeed, trees have specific algebraic properties that will be taken advantage of in Section \ref{sec_inference} for the inference of the model, but the model itself can be extended to other subsets of decomposable graphs.

\cite{Byrne2015} defined an extension of the (hyper) Markov properties described in \cite{Dawid1993} to undirected decomposable graphs (and to directed acyclic graphs, but this will not be discussed here) called the structural Markov property.

Let $\mathfrak{U}$ be the set of undirected decomposable graphs on $V$. A pair of subsets $(A,B)$ of $V$ is called a covering pair if $A\cup B = V$. For any family of graphs $\G \subseteq \mathfrak{U}$ and for any covering pair $(A,B)$, we define $\G(A,B)$ to be the set of graphs $G \in \G$ for which $(A,B)$ is a decomposition.

\begin{definition}
A distribution $\xi$ for $G \in \mathfrak{U}$ is said to be structurally Markov if for any covering pair $(A,B)$ such that $\xi(\mathfrak{U}(A,B)) > 0$, $G_A \indep G_B | \{G \in \mathfrak{U}(A,B) \}$ under $\xi$.
\end{definition}
A graph family supporting a structurally Markov graph distribution has some intrinsic structural property. It satisfies the so called structural meta Markov property.
\begin{definition}
Let $\G$ be a family of undirected decomposable graphs on $V$. Then $\G$ is structurally meta Markov if for any covering pair $(A,B)$, the set $\{G_A | G \in \G(A,B),G_B = J\}$ is the same for all $J \in \{G_B | G \in \G(A,B)\}$.
\end{definition}

The set of spanning trees $\T$ is an example of such a family \citep[Ex. 3.1]{Byrne2015} and the distributions that we considered in Section \ref{subsec:prior_structure} are structurally Markov.

These graph distributions naturally interact with (strong) hyper Markov hyperdistributions and Markov distributions when they are chosen carefully. Compatible hyperdistribution families, as described in Proposition \ref{prop_compatible_family}, conjugate nicely with graph distributions factorised on the edges, so that all hyperdistribution updates can be performed down to the edge level. But compatibility can be defined for any structurally meta Markov family $\G$ \citep[Definition 3.4]{Byrne2015}. Then, the update can be performed locally on $\mathcal{C}_{\G} = \bigcup_{G\in \G} \mathcal{C}_{G}$ where $\mathcal{C}_{G}$ denotes the cliques of graph $G$.

We finish this section by laying stress upon the fact that, among structurally meta Markov graph families, trees are of particular computational interest given their algebraic properties. One of the main difficulties in assessing graph distributions is to compute normalising constants, but closed-form expressions can be derived for these constants in the case of trees (see Section \ref{subsec_matrixtree}).

%% file: tex/inference.tex
\section{Inference in tree graphical models}
\label{sec_inference}

Different inference tasks can be performed on graphical models. One might be interested in estimating the emission distribution of $X$. \cite{Chow68} described an algorithm that can be used to get the tree distribution maximizing the likelihood of discrete multivariate data in the frequentist equivalent of the model given in the previous section. It can easily be adapted to MAP estimation in a full Bayesian framework \citep{Meila1999}. It is also possible to look at the posterior predictive distribution $p(\x|D) = \sum_{T\in \T} p(\x|T)\xi(T|D)$ \citep{Meila06}. In some other situations, the dependence structure between the variables, that is the graph $G$, might be the only object of interest. \cite{Lin2009} were for instance interested in the probability of an edge appearing in a tree. They looked out for the matrix $\beta$ maximising the likelihood of the data under a mixture of all possible tree models, where the probability of a tree model is defined just as in (\ref{prior_structure}). In their approach, the parameters of the models are estimated with plug-in estimators. Even if the distribution on trees cannot be called a prior in the traditional sense, the likeness to the model that we have described is obvious.

Here we are also interested in the probability for edges to appear in a tree, but in a full Bayesian framework. Formally, we would like to compute, for any edge $\{k,l\}$,
  \begin{align}
  \label{edge_prob}
 P(\{k,l\} \in E_{T^*} |D,\xi) = \sum_{T\in \T: E_T \ni \{k,l\}}\xi(T|D).
 \end{align}
 The previous section shows that achieving this requires two things. First, we have to get access to $\omega$ by computing local marginal likelihoods, which amounts to integrating w.r.t. $\pi$ (Section \ref{subsec_pi_int}). Then comes in the integration over the set of trees, that can be performed exactly using an algebraic result called the Matrix-Tree theorem (Section \ref{subsec_matrixtree}).

\subsection{Integration with respect to $\pi$}
\label{subsec_pi_int}

Thanks to the strong hyper Markov property required for the hyperdistributions, the integration on $\pi$ can be performed locally and the compatibility ensures that these local integrated quantities can be passed from one tree graphical model to another whenever they are needed. Thus, the integrations are always carried out on sets of bivariate distributions, with $p(p+1)/2$ of them to be computed. The small dimension of each of the involved problems makes it possible to consider numerical or Monte Carlo integration. We begin by describing a framework based on tree-structured copulas where it might be needed, depending on the choice of local copulas. We then present two settings where the local integrated likelihood terms can be computed exactly by using conjugate priors for the local distributions.

\subsubsection{Tree-Structured Copulas}

Let us assume that $\mathcal{X} = [0,1]^p$. If we make the assumption that the marginal distribution of each variable is uniform, the joint distribution for $\X$ is called a copula. Here we are interested in a subset of these distributions called the tree-structured copulas \citep{Kirshner2008}. We let $\UU$ denote the uniform distribution on $[0,1]$ and we assume that, for all $i\in V$, $X_i \sim \UU$. We are basically considering a copula model where the marginal data distributions have been dealt with in a relevant manner, independently from our model. For any $i \in V$, the marginal hyperdistribution $\rho_i$ for $\pi_i$ is then a Dirac distribution concentrated on $\UU$, denoted by $\delta_{\UU}$. Defining a compatible family of hyperdistributions requires that we consider pairwise hyperdistributions with marginals equal to $\delta_{\UU}$. Such hyperdistributions are in fact defined on bivariate copulas.

As an example, we consider the particular class of Archimedean copulas \citep{Nelsen2006}. The cumulative distribution function (cdf) of such copulas admit a simple expression. Let $\psi : [0,1] \rightarrow \R^+\cup\{\infty\}$ be a continuous, strictly decreasing function such that $\psi(1) = 0$. Its pseudo-inverse $\psi^{[-1]} : \R^+\cup\{\infty\} \rightarrow [0,1]$ is the continuous function defined by
\begin{align*}
\forall t\in \R^+\cup\{\infty\},~\psi^{[-1]}(t) = \left\lbrace\begin{array}{ll}
\psi^{-1}(t) & \textrm{if}~0\leq t \leq \psi(0), \\ 
0 & \textrm{otherwise}.
\end{array}\right.
\end{align*}
Let us remark that if $\psi(0) = \infty$, $\psi^{[-1]} = \psi^{-1}$. The cdf of the Archimedean copula generated by $\psi$ is given by $C_{\psi}(x_i,x_j) = \psi^{[-1]}(\psi(x_i) + \psi(x_j))$. Function $\psi$ is said to be a generator of the copula $C_{\psi}$. There is an extensive list of commonly used families of generators, many of them being governed by one or more parameters. Once again, we refer the reader to \cite{Nelsen2006} for a detailed list of such generators. We can mention the well-known Gumbel copulas as an example. 

Let $\{i,j\}$ be a given edge. If we consider an identifiable parametric family of Archimedean copulas $\{C_{\theta}\}_{\theta \in \Theta}$, $\Theta \subseteq \R$, defined by parametric generators $\{\psi_{\theta}\}_{\theta \in \Theta}$, there is a one-to-one mapping $\Upsilon$ between $\theta$ and the distributions $\pi_{ij}$ on $(X_i,X_j)$. A pairwise hyperdistribution $\rho_{ij}$ for $\pi_{ij}$ is then defined by any distribution $\kappa$ for $\theta$ through the identity $\rho_{ij}(\pi_{ij}) = \kappa\left(\Upsilon^{-1}(\pi_{ij})\right)$ and the integrated pairwise distribution $p(x_i,x_j)$ is given by
\begin{align}
\label{copula_likelihood}
p(x_i,x_j) &= \int_{\Theta} \frac{\partial^2 C_\theta}{\partial x_i \partial x_j} (x_i,x_j)\kappa(\theta)d\theta, & \forall (x_i,x_j) \in [0,1]^2.
\end{align}
Such a family of pairwise hyperdistributions is bound to be consistent since all marginals are equal to $\delta_{\UU}$. Morever, the global hyperdistributions that we obtain from this family are strong hyper Markov since it holds that, for $i,j \in V$, $\pi_{i|j} \indep \pi_{j}$ under $\rho_{ij}$.

The integrals given in (\ref{copula_likelihood}) shall be computed exactly or through numerical integration depending on the choice of the copula family. This choice needs not be the same for all the edges. In the case of Gumbel copulas, a numerical or Monte Carlo integration is required. Bivariate Gaussian copulas would also be a valid choice. The pairwise hyperdistributions could then be specified through Wishart distributions for the precision matrices of the copulas, just like in the full Gaussian case described in Section \ref{subsubsec_gaussian}.

\subsubsection{Multinomial Distributions}
\label{subsubsec_multinomial}

We now consider the case where all $X_i$ are discrete, taking values in finite spaces $\mathcal{X}_i$ of size $r_i$ respectively. Let $\mathcal{X}$ be the Cartesian product of spaces $\mathcal{X}_i$. A distribution for $\X$ is given by a probability vector $\theta$ in 
\begin{align*}
\Theta = \Set{\theta \in [0;1]^{\mid \mathcal{X} \mid} | \sum_{\x\in \mathcal{X}}\theta(\x) = 1 }.
\end{align*}
This is the set of multinomial distributions on $\mathcal{X}$. It happens that the conjugate Dirichlet distribution is satisfying the condition given in (\ref{cond_comp}) necessary to build a compatible family of strong hyper Markov hyperdistributions. 
Let $\lambda = (\lambda(\x))_{\x\in \mathcal{X}}$ be a family of positive numbers indexed by $\mathcal{X}$. For $\theta \in \Theta$, we let $\D(\lambda)$ denote the Dirichlet distribution, with density$f(\theta | \lambda) \propto \prod_{\x\in \mathcal{X}}\theta(\x)^{\lambda(\x)-1}$.
\begin{proposition}{\citep[Lemma 7.2]{Dawid1993}}
\label{prop_dirichlet} Let $A \subseteq V$ and $B = V \setminus A$. For all $\x_A \in \mathcal{X}_A$, we define $\lambda_{A}(\x_{A}) \defeq \sum_{\mathbf{y}, \mathbf{y}_{A} = \x_{A}}\lambda(\mathbf{y})$. If $\theta \sim \D(\lambda)$, then  $\theta_{A} \sim \D(\lambda_{A})$ and $\theta_{A} \indep \theta_{B|A}$.
\end{proposition}
It results from the fact that, if $\{Y_k\}_{k=1}^K$ are independent random variables distributed as $\Gamma(\lambda_k,\theta)$ respectively and if  $V \defeq \sum_{k=1}^K Y_k$, then $(Y_1/V,...,Y_K/V) \sim \D(\lambda)$. Proposition \ref{prop_dirichlet} states that any $\lambda$ gives rise to a hyperdistribution $\rho$ on the multinomial family of distributions from which we can build a family of compatible strong hyper Markov hyperdistributions and that the marginal hyperdistributions are also Dirichlet distributed. The conjugacy can then be used locally to compute $\omega$. These hyperdistributions were referred to as hyper-Dirichlet laws in \cite[\S 7.2.2]{Dawid1993}.

 As mentioned in Section \ref{subsec_prior_param}, specifying a full set of hyperparameters $\lambda$ is in fact not necessary to define the family of hyperdistributions $\{\rho^T\}_{T\in\T}$. We only need a consistent family of $\{\lambda_{ij}\}_{(i,j)\in V^2}$, in the sense that, for $(i,j,k) \in V^3$, $\lambda_{ij}$ and $\lambda_{ik}$ should induce the same $\lambda_i$. A possibility is to use an equivalent sample size $N$ and to set, for all $(i,j)\in V^2$, $\lambda_{ij} \defeq N/r_i r_j$ and $\lambda_{i} \defeq N/r_i$.
 If all $\mathcal{X}_i$ are of equal size $r$, one can choose $N = r^2/2$ so that all $\lambda_{ij}$ are equal to $1/2$ to mimic Jeffreys priors for the bivariate distributions on the edges. However, this choice will not induce global Jeffreys priors, which do not belong to hyper-Dirichlet hyperdistributions \citep{York1992}. For an edge $\{i,j\}$, we let $\lambda'_{ij}$ denote the updated hyperparameters for the edge $\{i,j\}$ given by $\lambda'_{ij}(\ell,\ell') = \lambda_{ij}(\ell,\ell') + \sum_{k=1}^n \delta_{x^{k}_i,\ell}\delta_{x^{k}_j,\ell'},~ \forall(\ell,\ell') \in \mathcal{X}_i\times \mathcal{X}_j$, where $\delta_{x,\ell} = 1$ if $x = \ell$ and $0$ otherwise. The matrix $\omega$ defined in (\ref{omega}) is then given by \citep{Meila06}
\begin{align}
\label{omega_multinomial}
\omega_{ij} =& \beta_{ij} \prod_{\ell \in \mathcal{X}_i}\frac{\Gamma(\lambda_{i}(\ell))}{\Gamma(\lambda'_{i}(\ell))}\prod_{\ell' \in \mathcal{X}_j}\frac{\Gamma(\lambda_{j}(\ell'))}{\Gamma(\lambda'_{j}(\ell'))}\prod_{(\ell,\ell') \in \mathcal{X}_i\times \mathcal{X}_j}\frac{\Gamma(\lambda'_{i,j}(\ell,\ell'))}{\Gamma(\lambda_{i,j}(\ell,\ell'))} \nonumber
\end{align}
where $\Gamma$ denotes the gamma function. If $R = \max_{i \in V}r_i$, computing $\omega$ requires $O(np^2R^2)$ operations \citep{Meila06}.

Let us finish this section by a remark on parameter independence. The following property of the Dirichlet distribution can be added to Proposition \ref{prop_dirichlet}.

\begin{proposition}{\citep[Lemma 7.2]{Dawid1993}}
Let $\theta \sim \D(\lambda)$. Then for all $A \subseteq V$ and $B = V \setminus A$, $\theta_{B|A}(\cdot | \x_{A})$ are all independent and distributed as $\D(\lambda_{B|A}(.|\x_{A}))$ with $\lambda_{B|A}(\x_{B}|\x_{A}) = \lambda(\x)$ for all $\x \in \mathcal{X}$ (up to a rearrangement of the components of $\x$).
\end{proposition}
Thus, although not required here, the local independence assumption made by \cite{Meila06} is in fact satisfied.  In the multinomial case, \cite{Geiger1997} even showed that, together with likelihood equivalence, global parameter independence and parameter modularity,  the local parameter independence assumption constrains the prior to be locally Dirichlet distributed.

\subsubsection{Gaussian Distributions}
\label{subsubsec_gaussian}

Whenever $\X$ is real-valued, one might work under the assumption that $\X$ is Gaussian-distributed with mean $\mu$ and inverse covariance matrix $\Lambda$. The conjugate normal-Wishart distribution is then a natural choice of prior for $(\mu, \Lambda)$. We let $n\mathcal{W}(\nu,\lambda,\alpha,\Phi)$ denote the normal-Wishart distribution hierarchically defined by
\begin{align*}
\Lambda  \sim \mathcal{W}(\alpha,\Phi),& &
 \mu|\Lambda  \sim \mathcal{N}(\nu,(\lambda\Lambda)^{-1}),
\end{align*} 
where $\mathcal{W}(\alpha,\Phi)$ stands for the Wishart distribution with $\alpha > p-1$ degrees of freedom and  positive-definite parametric matrix $\Phi$. \cite{Geiger02} showed that the normal-Wishart distribution satisfies the parameter independence property given in (\ref{cond_comp}). They further proved that this property coerces the distribution to be normal-Wishart whenever $p \geq 3$. It can thus be used to build a compatible family of strong hyper Markov hyperdistributions. Moreover, for any partitioning $(A,B)$ of $V$, $\X_A \sim \mathcal{N}(\mu_A,\left(\Lambda_A - \Lambda_{AB}\Lambda^{-1}_B\Lambda_{AB}^T\right)^{-1})$ and $(\mu_A,\Lambda_A - \Lambda_{AB}\Lambda^{-1}_B\Lambda_{AB}^T)$ is also normal-Wishart-distributed with parameters $(\nu_A,\lambda,\alpha-p+l,\Phi_A - \Phi_{AB}\Phi^{-1}_B \Phi_{AB}^T)$ where all indices are understood as partitioning of the corresponding vectors and matrices according to $(A,B)$.

The pairwise marginal likelihoods can then be computed by updating the hyperparameters of the basis hyperdistribution to $(\nu',\lambda',\alpha',\Phi')$, applying classical Bayesian updating formul\ae. The locally updated hyperparameters are then derived from the globally updated ones and 
\begin{align}
p(D_i,D_j)  \propto \frac{| \Phi_{\{i,j\}} |^{\frac{\alpha -p + 2}{2}}}{| \Phi'_{\{i,j\}} |^{\frac{\alpha' -p + 2}{2}}},\hspace{0.5cm} &
p(D_i)  \propto \frac{| \Phi_{i} |^{\frac{\alpha -p + 1}{2}}}{| \Phi'_{i} |^{\frac{\alpha' -p + 1}{2}}} ,
\end{align}
where, for a matrix $M$ and $i,j \in V$, $M_{\{i,j\}}$ denotes the submatrix of size 2 corresponding to vertices $i$ and $j$. This result is given in the work of \cite{Kuipers2014} as a correction to the erroneous result stated in \cite{Geiger02}.

The compatible hyperdistributions built on $(\mu,\Lambda)$ are called hyper-normal-Wishart distributions. One can notice that $\Lambda^{-1}$ follows a hyper-inverse-Wishart distribution \citep[\S 7.3.2]{Dawid1993}.


\subsection{Integration with respect to $T$}
\label{subsec_matrixtree}

We assume that we have knowledge of $\omega$. Consequently, we know $\xi(\cdot|D)$ up to the normalising constant $Z$. For an edge $\{k,l\}$, gaining access to $P(\{k,l\} \in E_{T^*} |D,\xi)$ means being able to sum the posterior tree distribution over the trees that borrow edge $\{k,l\}$. Because we are only considering trees, these summations can be efficiently performed.

Let $\omega = (\omega_{ij})_{(i,j)\in V^2}$ be a symmetric weight matrix such that, for all $i\in V$, $\omega_{ii} = 0$, and with non-negative off-diagonal terms. The weight of a graph $G=(V,E_G)$ is defined as the product of the weights of its edges, $\omega_G \defeq \prod_{\{i,j\} \in E_G} \omega_{ij}$. The Laplacian $\Delta=(\Delta_{ij})_{(i,j)\in V^2}$ of $\omega$ is given by $\Delta_{ij} = -\omega_{ij}$ if $i\ne j$ and $\Delta_{ii} = \sum_{j\in V}\omega_{ij}$ for $i\in V$.
For $U\subseteq V$, we defined $\Delta^{U}$ as the matrix obtained from $\Delta$ by removing the rows and columns corresponding to $U$, \LS{}{with rows and columns indexed by $V\setminus U$.}

\begin{theorem}[\citealp{Chaiken1982}]
\label{th:MT}
Let $\Delta$ be the Laplacian of a weight matrix $\omega$. Then all minors $|\Delta^{\{u\}}|$,  $u\in V$, are equal and $|\Delta^{\{u\}}| = \sum_{T\in \T} \omega_{T}$.
\end{theorem}
We directly get the normalising constant of $\xi(T|D)$ from this result.\

There is a more general version of this theorem concerning graphs whose connected components are spanning trees on their respective sets of vertices. Such graphs are called forests.

\begin{theorem}[All Minors Matrix-Tree theorem, \citealp{Chaiken1982}]
\label{allminorMT}
Let $\Delta$ be the Laplacian of a weight matrix $\omega$ and $U\subseteq V$. Let $\F_U$ be the set of forests on $V$ with $|U|$ connected components such that, for any two vertices $u_1,u_2 \in U$, $u_1$ and $u_2$ are not in the same connected component. Then $|\Delta^{U}| = \sum_{F\in \F_U}\omega_{F}$.
\end{theorem}
Briefly speaking, $U$ can be seen as a set of ``roots" (even though the models are not directed) for the trees of the forests in $\F_U$. If $U$ is taken equal to a single vertex, then the forests in $F_U$ only have one connected component which is a tree and we get Theorem \ref{th:MT}. This theorem will be used in the proof of the following result that was first stated by \cite{Kirshner2008}.

\begin{theorem}[\citealp{Kirshner2008}]
\label{th:Kirshner}
Let $\omega$ be defined as in (\ref{omega}) and $\Delta$ be the associated  Laplacian. Let $u$ be a vertex in $V$. \LS{}{We define matrices $Q$ and $M$ respectively by
\begin{align}
Q_{kl}  & = \left\lbrace \begin{array}{ll}
\left[\left( \Delta^{\{u\}} \right)^{-1}\right]_{kl} & \textrm{if}~k,l \ne u, \\
0 & \textrm{otherwise},
\end{array} \right. \\ 
M_{kl} & = Q_{kk} + Q_{ll} - 2Q_{kl}. \label{eq:M}
\end{align}
Then, for all $\{k,l\}\in \mathcal{P}_2(V)$,
\begin{align}
P(\{k,l\} \in E_{T^*} | D,\xi) = \omega_{kl}\cdot M_{kl}
\end{align}}
\end{theorem}
A proof of this result is provided in the extended version of \citep{Kirshner2008} available online. We provide a shorter version relying on the generalized version of the Matrix-Tree theorem given above.
\begin{proof}
Let $\{k,l\}$ be an edge in $\mathcal{P}_2(V)$. Let $Z$, $Z_{kl}^+$ and $Z_{kl}^-$ respectively denote the sums of $\omega_T$ over the sets $\T$, $\{T \in \T : \{k,l\} \in E_T\}$ and $\{T \in \T : \{k,l\} \not\in E_T\}$. It is immediate to see that $Z = Z_{kl}^+ + Z_{kl}^-$. Lemma 3 of \citep{Meila06} states that $\frac{\partial Z}{\partial \omega_{kl}} = M_{kl}|\Delta^{\{u\}}| = M_{kl}Z$ \LS{}{where M is defined as in (\ref{eq:M}).}
It is easy to see that $Z_{kl}^-$ can be obtained by applying Theorem \ref{th:MT} to a weight matrix equal to $\omega$ except for the terms $\omega_{kl}$ and $\omega_{lk}$ that are set to $0$. This means that $Z_{kl}^-$ does not depend on $\omega_{kl}$ and $\frac{\partial Z}{\partial \omega_{kl}} = \frac{\partial Z_{kl}^+}{\partial \omega_{kl}}$.

We then use Theorem \ref{allminorMT} to get an expression of $Z_{kl}^+$. Indeed, there is a one-to-one correspondence between the set of forests rooted in $k$ and $l$ (denoted by $\F_{\{k,l\}}$) and  the set of trees borrowing edge $\{k,l\}$. Going from one to the other is just a matter of adding or removing edge $\{k,l\}$. Then, by Theorem \ref{allminorMT},
\begin{align}
\label{eq:Zkl+}
Z_{kl}^+ = \omega_{kl}\sum_{F\in \F_{\{k,l\}}} \omega_{F} = \omega_{kl} \cdot |\Delta^{\{k,l\}}|.
\end{align}
$|\Delta^{\{k,l\}}|$ does not depend on $\omega_{kl}$ since the only terms of $\Delta$ that depend on $\omega_{kl}$ are $\Delta_{kl}$, $\Delta_{lk}$, $\Delta_{kk}$, $\Delta_{ll}$ and these terms are all withdrawn in $\Delta^{\{k,l\}}$. Therefore,
\begin{align}
\label{eq:Delta_kl}
|\Delta^{\{k,l\}}| = \frac{\partial Z_{kl}^+}{\partial \omega_{kl}}  =\frac{\partial Z}{\partial \omega_{kl}} = M_{kl}\cdot Z.
\end{align}
Combining (\ref{eq:Zkl+}) and (\ref{eq:Delta_kl}) with the fact that $P(\{k,l\} \in E_T | D,\xi) = Z_{kl}^+/Z$, we get the claimed result.
\end{proof}
Theorem \ref{th:Kirshner} shows that posterior probabilities can be computed for all edges at once by inverting a matrix of size $p-1$, amounting to a total complexity of $O(p^3)$.\\

\subsection{Other quantities of interest}
\label{subsec_other_quantities}

\paragraph{{Posterior moments of the degree.}}
{
The aim of structure inference is to decipher the dependency structure of a set of random variable. In this perspective, the degree of vertex $k$ (i.e. its number of neighbors) in the graph informs us about the centrality of the corresponding variable $X_k$ in the system. Denoting $N_k$ this degree, we can easily derive the posterior mean of $N_k$ from the end of the proof of Theorem \ref{th:Kirshner} as
$$
\EE[N_k|D] = \sum_{l \neq k} Z^+_{kl} / Z \LS{}{ ~ = \sum_{l \neq k} P(\{k,l\} \in E_{T^*} | D,\xi)}
$$
}

\LS{}{The posterior variance of $N_k$ can also be computed for all vertices with total complexity $O(p^3)$. The proof of this result is based on the following lemma giving some the second-order derivatives of the normalising constant $Z$.

\begin{lemma}
\label{lemma:Zd2}
Let $\omega$ be defined as in (\ref{omega}) and $\Delta$ be the associated  Laplacian. Let $u$ be a vertex in $V$ and $Q$ defined as in Theorem \ref{th:Kirshner}. For $k\in V$, let $M^{(k)}$ be the matrix whose general term is given by
\begin{align*}
M^{(k)}_{l_1 l_2} & = M_{k l_1}M_{k l_2} - M_{k,l_1,l_2}^2,
\end{align*}
where $M_{l_1l_2}$ is defined as in (\ref{eq:M}) and $M_{k,l_1,l_2} \defeq  Q_{kk} + Q_{l_1l_2} - Q_{kl_1} - Q_{kl_2}$. Then, for $k, l_1, l_2 \in V$ that are pairwise distinct, it holds that
\begin{align*}
\frac{\partial^2 Z}{\partial\omega_{kl_1}\partial\omega_{kl_2}} = Z \cdot M^{(k)}_{l_1l_2}.
\end{align*}
\end{lemma}
The proof of this lemma is given in the Appendix.

\begin{theorem}
\label{th:var_deg}
Let $\omega$ be defined as in (\ref{omega}) and $\Delta$ be the associated  Laplacian. Let $u$ be a vertex in $V$ and $Q$ be defined as in Theorem \ref{th:Kirshner}. Then, for all $k\in V$, we let $\V(N_k|D)$ denote the posterior variance of  $N_k$ and it holds that
\begin{align}
\label{eq:var_Nk}
\V(N_k|D) &= \EE[N_k|D]\left(1 - \EE[N_k|D]\right) + \sum_{\substack{ l_1 \ne k, l_2 \ne k \\ l_1 \ne l_2 }} \omega_{k,l_1}\omega_{k,l_2}M^{(k)}_{l_1,l_2}.
\end{align}
\end{theorem}

\begin{proof}
We have that
\begin{align*}
\EE[N_k^2|D] = \sum_{\substack{l_1 \ne k\\ l_2 \ne k}}\EE[\mathbf{1}_{\{k,l_1\}}\mathbf{1}_{\{k,l_2\}}|D].
\end{align*}
Let $l_1,l_2 \in V\setminus \{k\}$ such that $l_1 \ne l_2$. There is a one-to-one correspondence  between the set of trees borrowing edges $\{k,l_1\}$ and $\{k,l_2\}$, and the forests rooted in $\{k,l_1,l_2\}$. Using Theorem \ref{allminorMT} and Lemma \ref{lemma:Zd2}, we deduce that 
\begin{align*}
\EE[\mathbf{1}_{\{k,l_1\}}\mathbf{1}_{\{k,l_2\}}|D] =  \omega_{k,l_1}\omega_{k,l_2}M^{(k)}_{l_1,l_2}
\end{align*}
by a reasoning similar to the one used in the proof of Theorem \ref{th:Kirshner}. The expression given in (\ref{eq:var_Nk}) is then easily derived.
%
\end{proof}

Theorem \ref{th:var_deg} shows that the posterior variance for the degree of all vertices can be obtain directly at virtually no extra cost once posterior edge probabilities have been computed, since both computations rely on the inversion of the same matrix.
}

\paragraph{{Posterior entropy.}}
In a Bayesian framework, the posterior entropy gives insight about the concentration of the posterior distribution, which is for instance of particular interest when a MAP approach is considered. The computation of this quantity is not always straightforward, but here, it can be obtained at small cost once posterior probabilities for the edges have been computed.

\begin{proposition}
The entropy of the posterior distribution on trees $\xi(\cdot|D)$ can be computed with complexity $O(p^3)$.
\end{proposition}
\begin{proof}
We show that the entropy has a simple expression depending on $Z$ and $( P(\{k,l\} \in E_{T^*} | D,\xi) )_{\{k,l\} \in \mathcal{P}_2(V)}$ which can both be computed with complexity $O(p^3)$ through Theorems \ref{th:MT} \& \ref{th:Kirshner}. Indeed,
\vspace{-0.2cm}
\begin{align*}
H(\xi(\cdot|D)) &= -\sum_{T\in \T}\xi(T|\D)\log\left(\xi(T|D)\right)   \\
 &= \sum_{T\in \T}\frac{1}{Z}\prod_{\{i,j\} \in E_T}\omega_{ij}\left(\log(Z) - \sum_{\{k,l\} \in E_T} \log(\omega_{kl})\right) \\
 & = \log(Z) - \sum_{\{k,l\}  \in \mathcal{P}_2(V)}\frac{\log(\omega_{kl})}{Z}\sum_{T\ni \{k,l\}}\prod_{\{i,j\} \in E_T}\omega_{ij} \\
 &= \log(Z)  - \sum_{\{k,l\} \in \mathcal{P}_2(V)} \log(\omega_{kl})P(\{k,l\} \in E_{T^*}|D,\xi).  
\end{align*}
\end{proof}

\subsection{Controlling prior edge probability}
\label{subsec_change_of_prior}

If the distribution on trees is not strongly peaked, the prior probability for an edge to appear in a random tree can be quite small. For instance, the uniform distribution on $\T$ leads to any edge appearing with probability $2/p$. Indeed, no edge is favoured and each tree borrows $p-1$ of the $p(p-1)/2$ possible edges. We consider an edge $\{k,l\} \in \mathcal{P}_2(V)$ and the event $\E_{kl} \defeq \left\lbrace T : \{k,l\} \in E_{T} \right\rbrace$. We let $p_{kl}^0$ and $p_{kl}$ respectively denote the prior and posterior probabilities of event $\E_{kl}$. These probabilities are obtained through Theorem \ref{th:Kirshner}. 

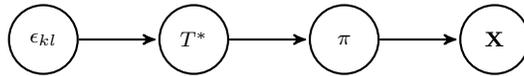
\begin{figure}[b]
\centering
\begin{tikzpicture}[->,>=stealth',shorten >=1pt,auto,node distance=2cm,
 		 thick,main node/.style={circle,fill=blue!0,draw,minimum size=26pt}];

		\node[main node] (1)  {\footnotesize $\epsilon_{kl}$};
		\node[main node] (2) [right of=1] {\footnotesize $T^*$};
		\node[main node] (3) [right of=2] {\footnotesize $\pi$};
		\node[main node] (4) [right of=3] {\footnotesize $\X$};

		\path[every node/.style={font=\sffamily\small}]
    		(1) edge node {} (2)
    		(2) edge node {} (3)
		(3) edge node {} (4);
		\end{tikzpicture}
		\caption{Model with variable $\epsilon_{kl}$ explicitly controlling the status of edge $\{k,l\}$ in $T^*$.}
		\label{fig:model_epsilon}
\end{figure}

In a decision perspective, it might be desirable to allow some control on the prior probability of $\E_{kl}$. To this aim, we use a binary random variable $\epsilon_{kl} \sim \mathcal{B}(\lambda_{kl})$ explicitly controlling the status of edge $\{k,l\}$ in the random tree:
\begin{align*}
p(T|\epsilon_{kl},\xi) = \left\lbrace\begin{array}{cc}
\xi(T|\E_{kl}) & \textrm{if}~\epsilon_{kl} = 1 \\ 
\xi(T|\overline{\E}_{kl}) & \textrm{if}~\epsilon_{kl} = 0
\end{array} \right. .
\end{align*}
In particular, the choice $\lambda_{kl} = 1/2$ takes us back to a non-informative prior configuration regarding $\E_{kl}$. We obtain the model represented in Figure \ref{fig:model_epsilon} in which the fully marginal likelihood can be written as
\begin{align*}
p(D) = \lambda_{kl} p(D|\E_{kl}) + (1 - \lambda_{kl}) p(D|\overline{\E}_{kl}).
\end{align*}

We are now interested in the posterior distribution of $\epsilon_{kl}$.

\begin{proposition}
\begin{align*}
&P(\epsilon_{kl} = 1|D) = \lambda_{kl}\frac{p_{kl}}{p_{kl}^0}\cdot\left[ \lambda_{kl}\frac{p_{kl}}{p_{kl}^0} + (1 - \lambda_{kl}) \frac{1 - p_{kl}}{1 - p_{kl}^0} \right]^{-1}
\end{align*}
\end{proposition}
\begin{proof}
\begin{align*}
P(\epsilon_{kl} = 1|D) &= \frac{p(D|\epsilon_{kl} = 1)P(\epsilon_{kl} = 1)}{p(D)} =  \lambda_{kl}\frac{p(D|\E_{kl})}{p(D)} \\
& = \lambda_{kl} p(D|\E_{kl})\cdot\left[\lambda_{kl} p(D|\E_{kl}) + (1 - \lambda_{kl})p(D|\overline{\E}_{kl})\right]^{-1} \\
& = \lambda_{kl}\frac{p_{kl}}{p_{kl}^0}\cdot\left[ \lambda_{kl}\frac{p_{kl}}{p_{kl}^0} + (1 - \lambda_{kl}) \frac{1 - p_{kl}}{1 - p_{kl}^0} \right]^{-1} 
\end{align*}
\end{proof}

The computation of $P(\epsilon_{kl} = 1|D) $ for all edges can be achieved in $O(p^2)$ time from the posterior edge probability matrix $\{p_{kl}\}_{\{k,l\} \in \mathcal{P}_2(V)}$. We can notice that $P(\epsilon_{kl} = 1|D)$ is a strictly increasing function of $p_{kl}$. When the initial prior on trees $\xi$ is uniform and all $\lambda_{kl}$ are taken equal,  the order induced on the edges by $\{P(\epsilon_{kl} = 1|D)\}_{\{k,l\} \in \mathcal{P}_2(V)}$ is identical to the order induced by the posterior edge probability matrix. The ROC and PR curves that are commonly used to assess network inference accuracy therefore remain unchanged.

%% file: tex/simuls.tex
\section{Simulations}
\label{sec_simu}

In this section, we use synthetic data to meet a twofold objective. On one hand, the aim of this study is to show that there is an advantage in averaging over trees rather than considering a single MAP estimate. On the other hand, we show that assuming a tree structure is not substantially more detrimental to the accuracy of the inference of non-tree-structured graphical models than assuming a DAG structure. To do so, we compare our method with another fully Bayesian inference method carried out on DAGs, described by \cite{NiinimakiPK11} and implemented in the BEANDisco software. Computations for our approach were performed with the \textsf{R} package \textbf{saturnin}.

 \begin{figure}[!b]
 \captionsetup[subfigure]{labelformat=empty}
 \subfloat[\textsf{Tree}]{\centering
 \includegraphics[width=0.33\linewidth]{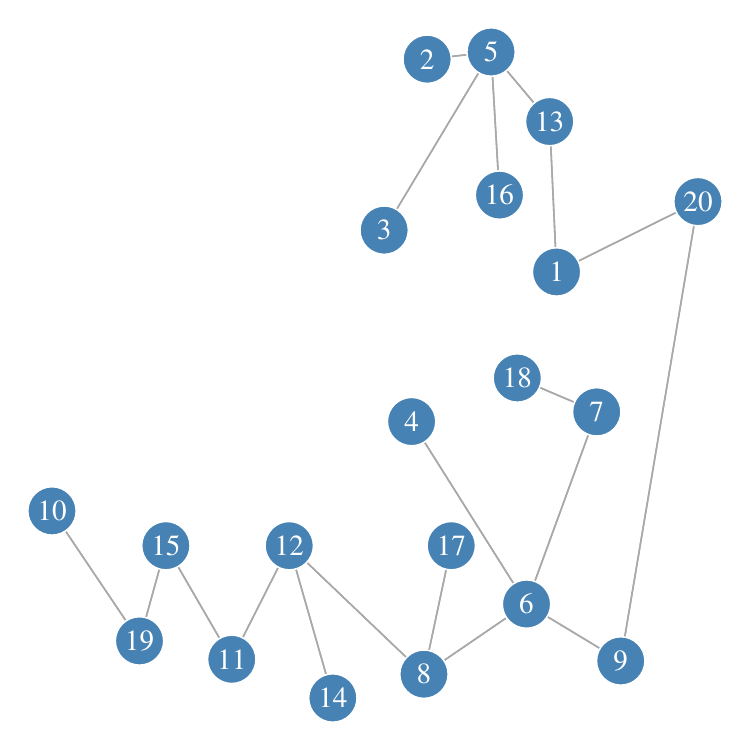}}
 \subfloat[\textsf{Non-tree, low density}]{\centering
 \includegraphics[width=0.33\linewidth]{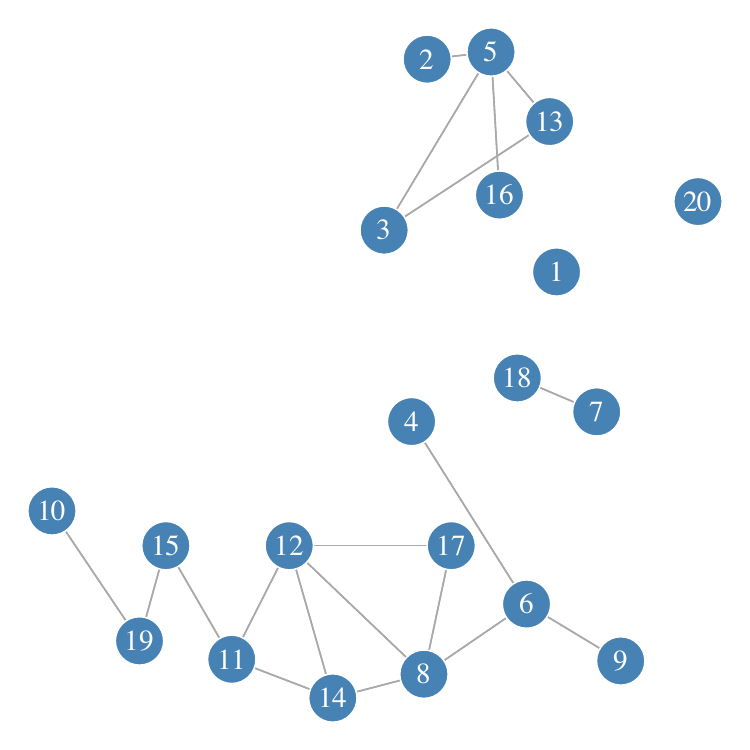}}
 \subfloat[\textsf{Non-tree, high density}]{\centering
 \includegraphics[width=0.33\linewidth]{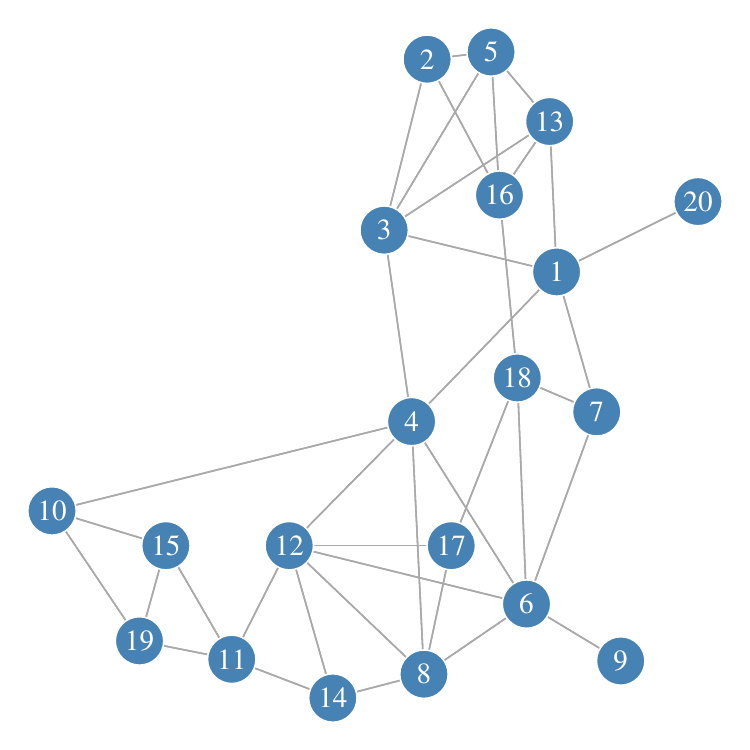}}
 \caption{Gold standard networks in the simulation study.}
     \label{fig_graph}
 \end{figure}

\subsection{Simulation scheme}

\LS{}{We have chosen three networks with $p = 20$ vertices. The first one is a spanning tree. The second and third graphs are not spanning trees and respectively have as many and twice as many edges as the first one. These graphs are shown in Figure \ref{fig_graph}. We then simulated data according to a multinomial model with $\mathcal{X}_i = \left\lbrace 1,2,3 \right\rbrace$ for $i \in V$. For each graph $G$, we have chosen a DAG $D$ with skeleton equal to $G$. We let $\textrm{par}(i,D)$ stands for the set made of the parents of vertex $i$ in DAG $D$.  For $\X \in \left\lbrace 1,2,3 \right\rbrace^p$, we let $N_i^D(r;\X) \defeq \big|\left\lbrace j \in \textrm{par}(i,D): X_j = r\right\rbrace \big| $ denote the number of parents of vertex $i$ in $D$ taking value $r \in \left\lbrace 1,2,3 \right\rbrace$ in $\X$. Then, conditionally on $D$, we used the following distribution for $\X$:
{
$p\left(X_i = r\right) = {1}/{3}$ if $\textrm{par}(i,D) = \varnothing$ and
\begin{align*}
p\left(X_i = r | X_{\textrm{par}(i,D)}\right) &\propto \eta \cdot N_i^D(r;\X) + 1 & \textrm{if}~\textrm{par}(i,D) \ne \varnothing.
\end{align*}
}
As the variables at root vertices are drawn uniformly, it can be shown that all vertices are marginally uniformly distributed by a symmetry argument. Here, $\eta$ was set to $0.5$. For $n=25$, 50, 75, 100 and 200,  we generated 100 samples of size $n$.}

We then considered the Multinomial/Dirichlet framework described in Section \ref{subsubsec_multinomial}, setting the prior on trees $\xi$ to the uniform and the equivalent prior sample size  $N$ to $3^2/2 = 4.5$ (see Section \ref{subsubsec_multinomial}). For each data set, we computed
\begin{itemize}
\item the MAP tree structure through a Maximal Spanning tree algorithm \citep{Prim1957} applied to $\omega$;
\item the matrix of posterior edge probabilities $P(\{k,l\} \in E_{T^*}|D)$ in our model. For all the edges, the prior appearance probability was set to $\lambda_{kl} = 1/2$ (see Section \ref{subsec_change_of_prior});
\item an estimation of the matrix of posterior edge appearance probabilities in a random DAG obtained by MCMC sampling \citep{NiinimakiPK11}. We refer the reader to this paper for details on the prior distribution on DAGs. We ran the code provided by the authors with default parameters. \LS{}{The sampling was performed for one minute on each dataset.} The direction of the edges of the sampled DAGs was not taken into account  to get empirical frequencies for all undirected edges.
\end{itemize}

The accuracy of the inference was evaluated against the true undirected adjacency matrix, according to the yielded outputs. In the case of the MAP estimate, we calculated the True and False Positives Rates (TPR, FPR) between the best tree and the true graph. These rates are constrained by the fact that a spanning trees on $p$ vertices has exactly $p-1$ edges. For the (estimated) posterior edge appearance probability matrices, ROC and PR curves against the true adjacency matrix are plotted and summarized by the area under the curves.

\subsection{Results}

\captionsetup{position=top}
 \begin{figure}[!p]
 \captionsetup[subfigure]{labelformat=empty} 
  
 \subfloat[\textsf{~~~~Tree}]{\centering
 \includegraphics[width=0.33\linewidth]{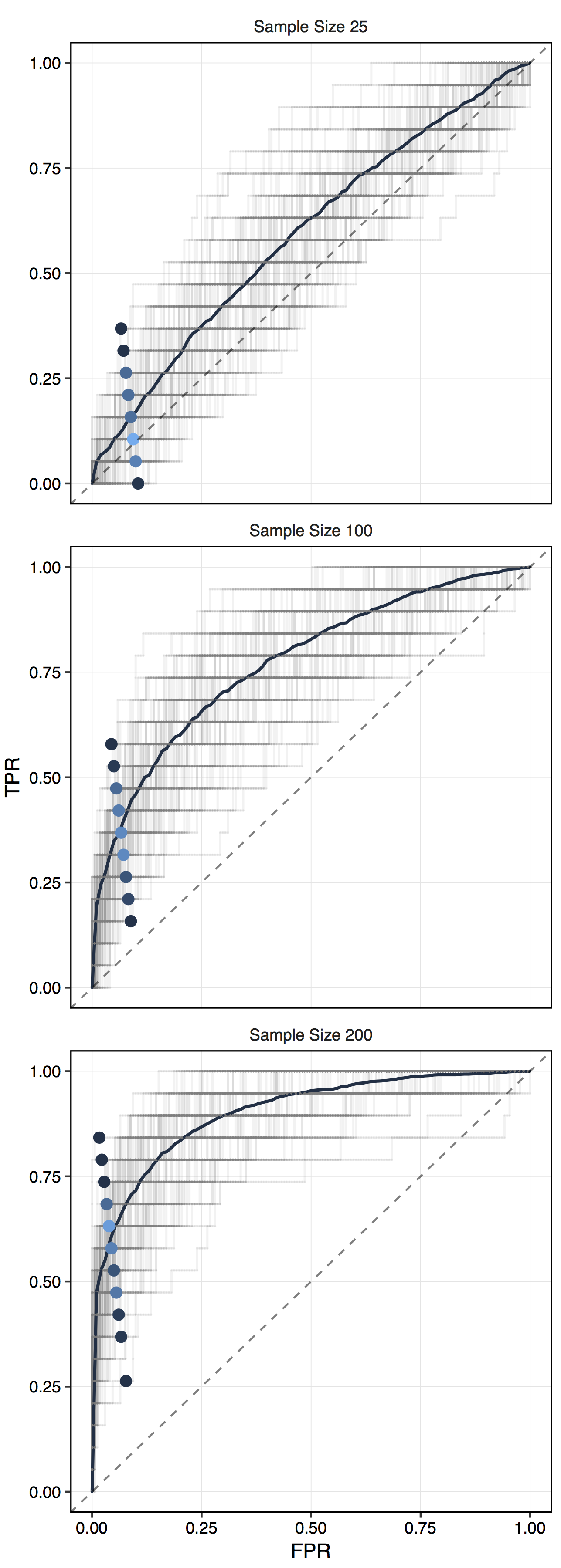}}
 \subfloat[\textsf{~~~~Non-tree, low density}]{\centering
 \includegraphics[width=0.33\linewidth]{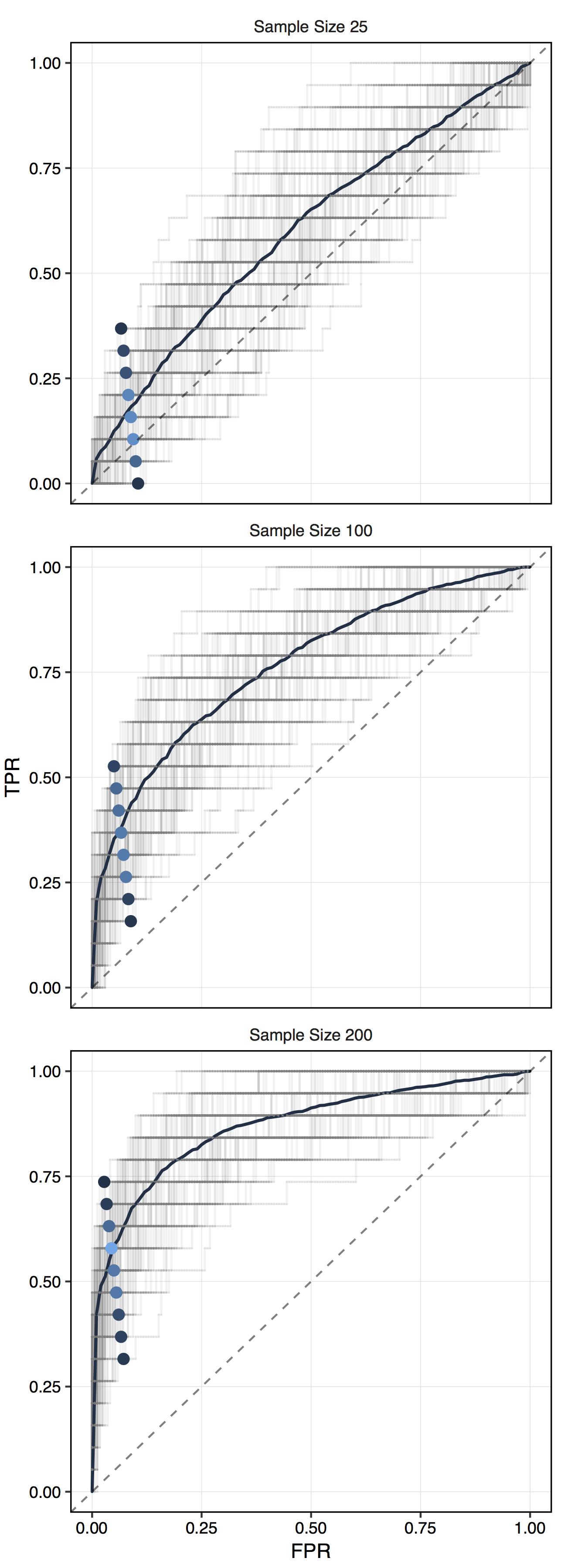}}
 \subfloat[\textsf{~~~~Non-tree, high density}]{\centering
 \includegraphics[width=0.33\linewidth]{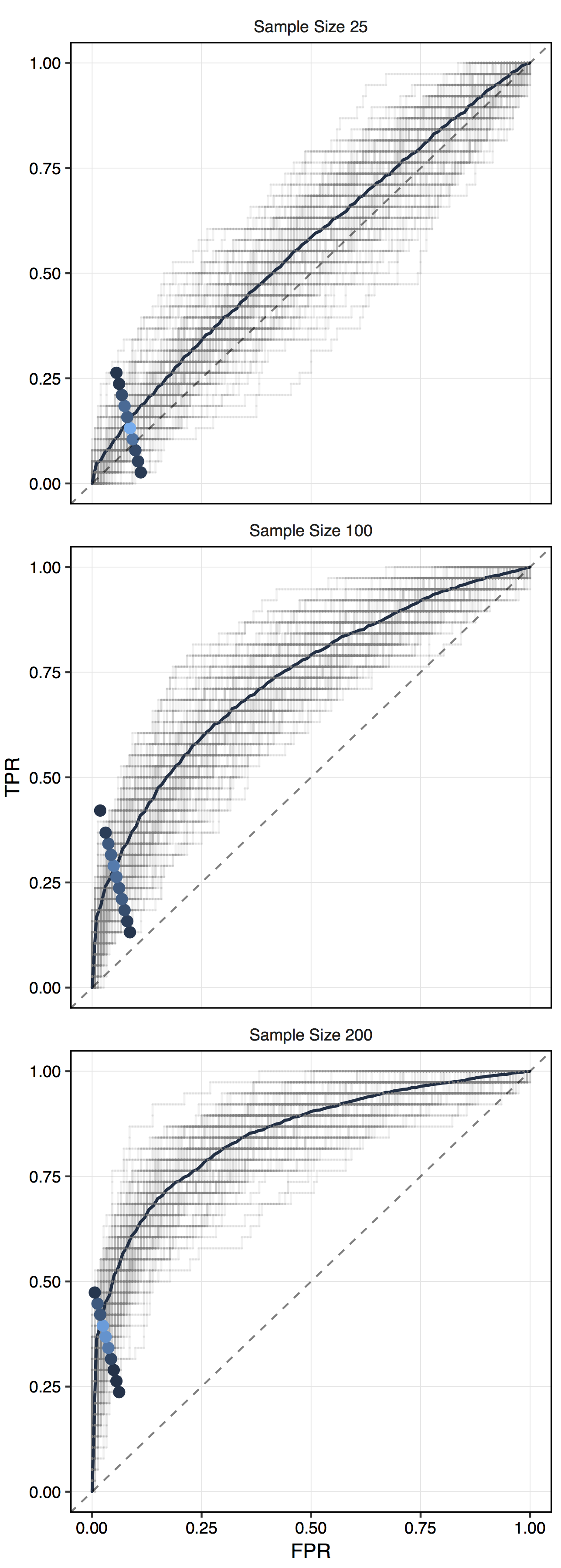}}
 
 \vspace{-5mm}
 
   ~~~~\includegraphics[width=30mm]{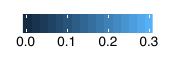}

 \caption{ROC curves for the posterior edge probabilities and (TPR, FPR) scores for the MAP estimate on data sets of size 25, 100 and 200 (from top to bottom). For the ROC curves, the mean curve is plotted in bold line. The color of a (TPR, FPR) point expresses its frequency within the 100 samples.}
 \label{fig_rocCL}
 \end{figure}


 \begin{figure}[p]
         \includegraphics[width=0.8\linewidth]{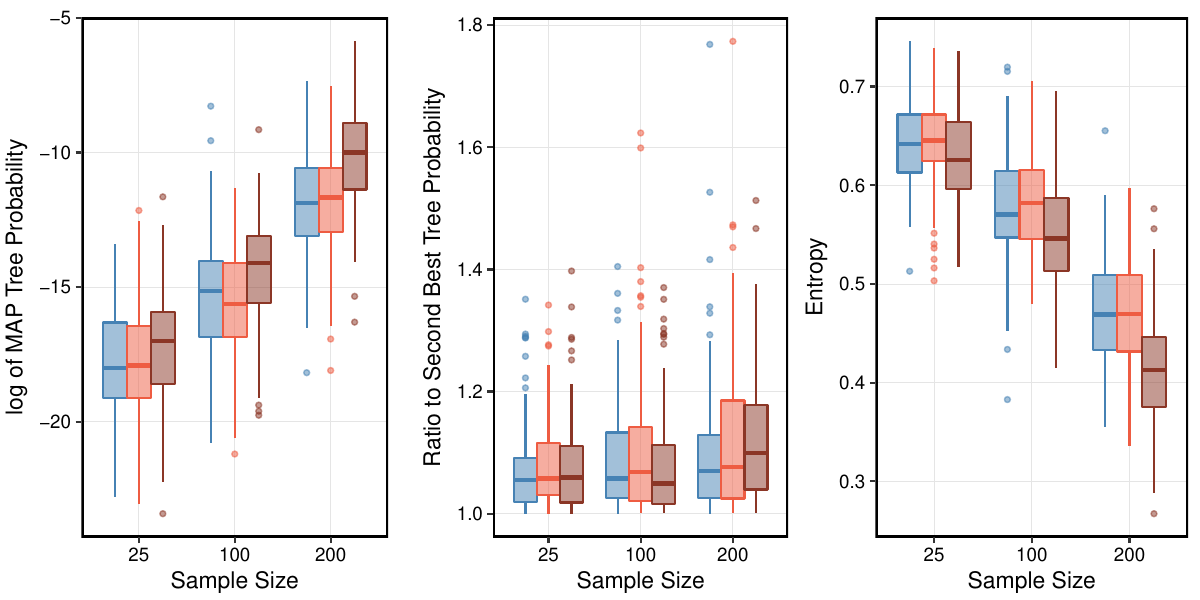}
         
         \crule[tree]{0.25cm}{0.25cm} \textsf{Tree} \hspace{1cm} \crule[ER2p]{0.25cm}{0.25cm} \textsf{Non-tree, low density}\hspace{1cm}
         \crule[ER8p]{0.25cm}{0.25cm} \textsf{Non-tree, high density}
 \caption{Posterior probability of the MAP tree, ratio to the posterior probability of the second best tree and entropy of the posterior tree distribution (normalised by the entropy of the uniform distribution on $\T$, \textit{i.e.} $(p-2)\log(p)$).}
 \label{secondbesttree}
 \end{figure}
 \begin{figure}[p]
 \captionsetup[subfigure]{labelformat=empty}   
 
 \subfloat[\textsf{~~~~Tree}]{\centering
 \includegraphics[width=0.33\linewidth]{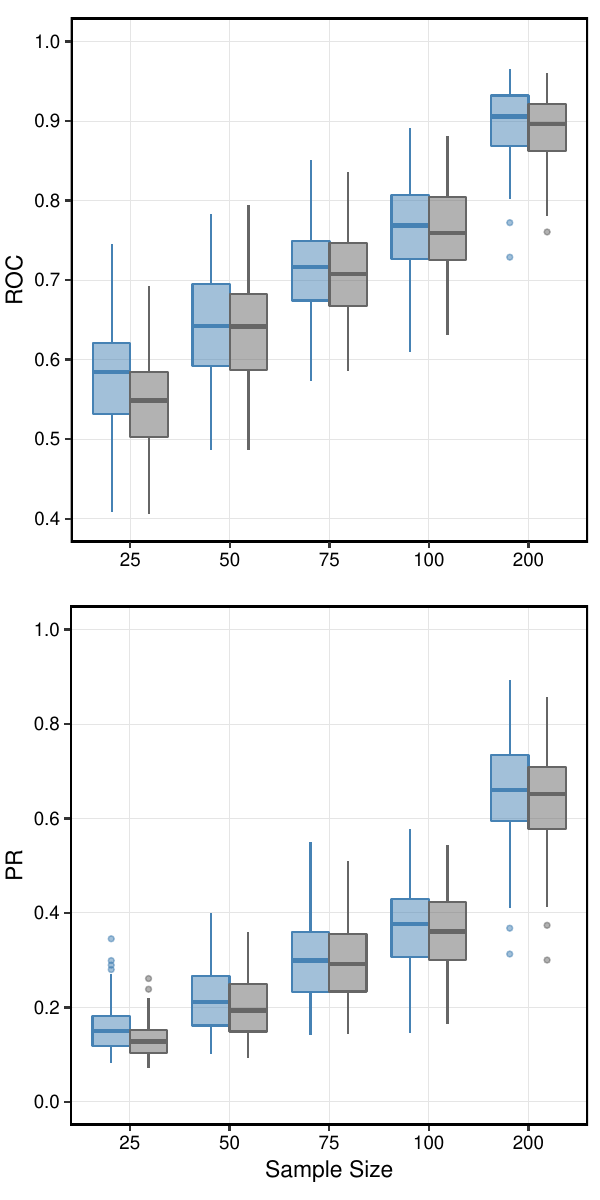}}
 \subfloat[\textsf{~~~~Non-tree, low density}]{\centering
 \includegraphics[width=0.33\linewidth]{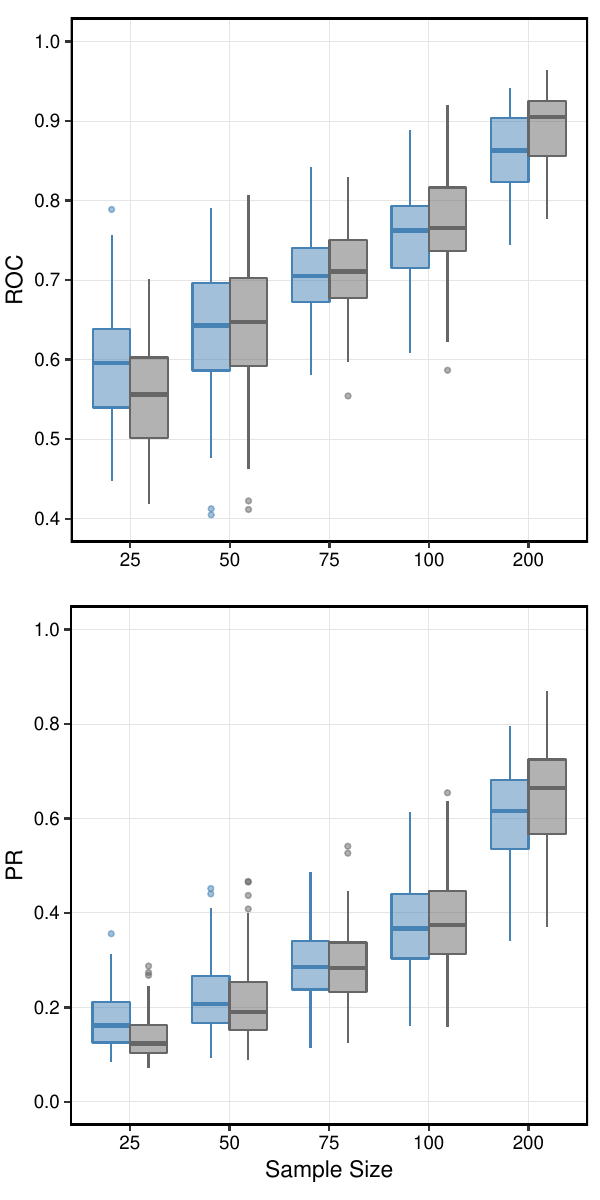}}
 \subfloat[\textsf{~~~~Non-tree, high density}]{\centering
 \includegraphics[width=0.33\linewidth]{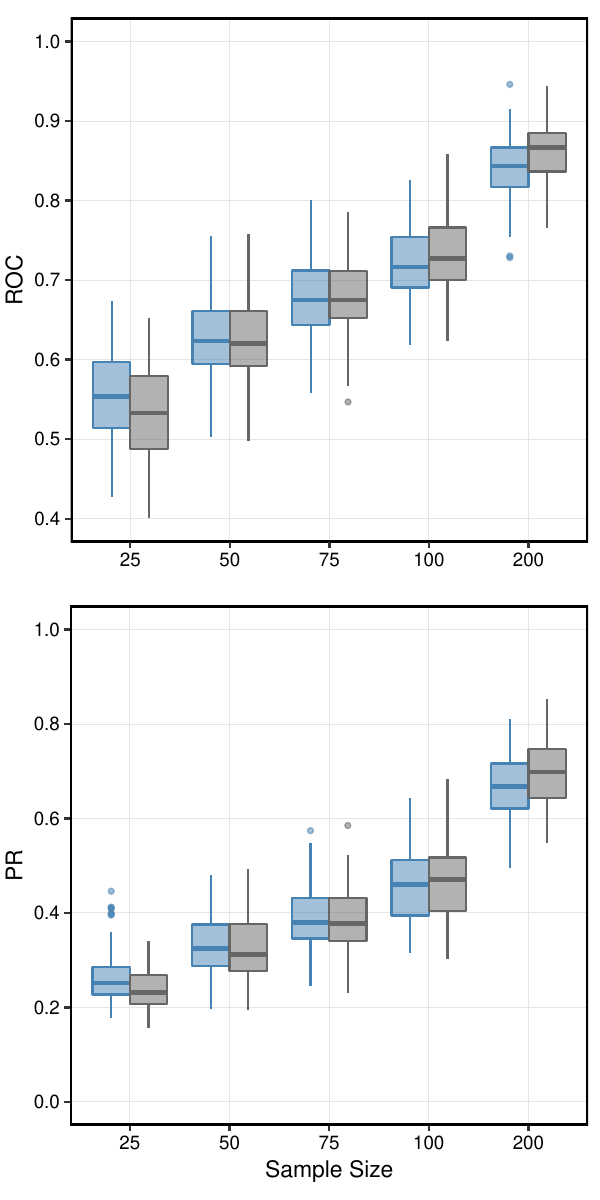}}
 
 \vspace{-4mm} 
 
   \begin{tabular}{clcl}
\crule[tree]{0.25cm}{0.25cm}  & \textsf{Tree} & \crule[gray!85]{0.25cm}{0.25cm} & \textsf{DAG} 
 \end{tabular} 

     \caption{Area under the ROC (top) \& PR (bottom) curves computed for the output of our approach and of the MCMC sampling algorithm in the set of DAGs.}
     \label{fig_simu}
 \end{figure}

\paragraph{Comparison with MAP}
Figure \ref{fig_rocCL} simultaneously represents the (TPR, FPR) scores and the ROC curves obtained for the MAP estimate and the tree posterior edge appearance probability matrix respectively. It makes sense to plot both results on the same graph since a ROC curve is just a succession of (TPR, FPR) points computed as more and more edges are selected, going from the most to the least likely. When $p-1$ edges are selected, both methods behave similarly. So, if there is external evidence that the true graph is in fact a tree, a MAP approach could be considered but using posterior edge probabilities would do as well. Nonetheless, when the true graph is not a tree, the MAP approach is penalised by its lack of flexibility. Computing posterior appearance probabilities for the edges allows to retain an arbitrary number of edges. The balance between selectivity and sensibility achieved by the MAP approach can obviously be improved by selecting more edges. An other argument in favour of considering the whole posterior distribution on trees instead of the MAP is presented in Figure \ref{secondbesttree}.  For all three simulation scenarios, posterior tree distributions are not really peaked around their modes, especially for small samples. The second most probable tree is always very close to the MAP. Moreover, the entropy of the posterior distribution on trees behaves similarly across all simulation scenarios.

\paragraph{Influence of the tree assumption} We now study the influence of the tree assumption on the accuracy of structure inference when the true graphical model is not tree-structured. With this end in view, we consider a similar model where DAGs are drawn instead of trees and use the posterior edge appearance probabilities yielded by this model as gold standard, as it achieves the same goal in terms of Bayesian inference within a larger class of graphs. Results are given in Figure \ref{fig_simu}. Both algorithms seem to perform equally well in all three scenarios. The accuracy of the inference expectedly increases with sample size. The results we get here indicate that the posterior probabilities for the edges to belong to a random tree can be relevant even when the true network is not a tree, with no clear evidence in favour of considering an inference within the broader class of DAG structures.\\

%

%% file: tex/appli.tex
\section{Application to cytometry data}
\label{sec_app}

\begin{figure}[!b]
         \includegraphics[trim = 12mm 35mm 12mm 18mm, clip,width=\linewidth]{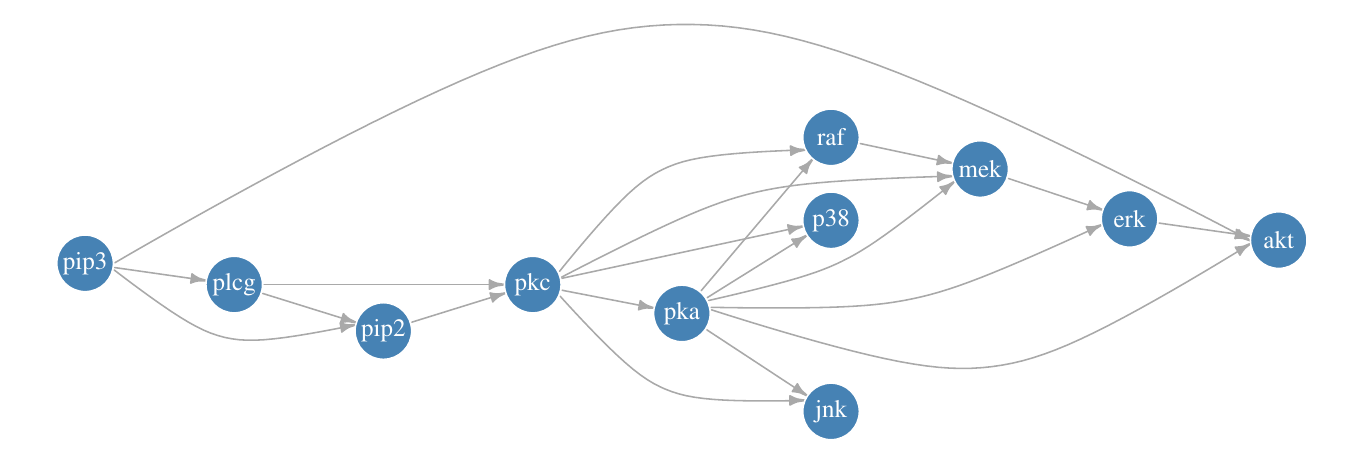}
 \caption{Raf pathway.}
 \label{RAF_network}
 \end{figure}

This section presents an application of our approach to flow cytometry data. They have been collected by \cite{Sachs2005} and were used by \cite{Werhli2006} in a review of
 network inference techniques. They are related to the Raf cellular signalling network, which is involved in many different biological processes, including the regulation of cellular proliferation in human immune cells. The activation levels of the 11 proteins and phospholipids that are part of this pathway were measured by flow cytometry. The generally accepted structure of the Raf pathway is given in Figure \ref{RAF_network}, but the true structure of this network, despite considerable experimental and theoretical efforts, may be more subtle. The undirected skeleton of this network will, however, be used as the gold standard network in our study.
 
 \captionsetup{position=bottom}

 \begin{figure}[!b]
 \vspace{-0.5cm}
 \subfloat[Most likely (left) and second most likely (right) trees in the posterior distribution on trees.\label{RAF_MAP}]{\centering
 \includegraphics[width=0.5\linewidth]{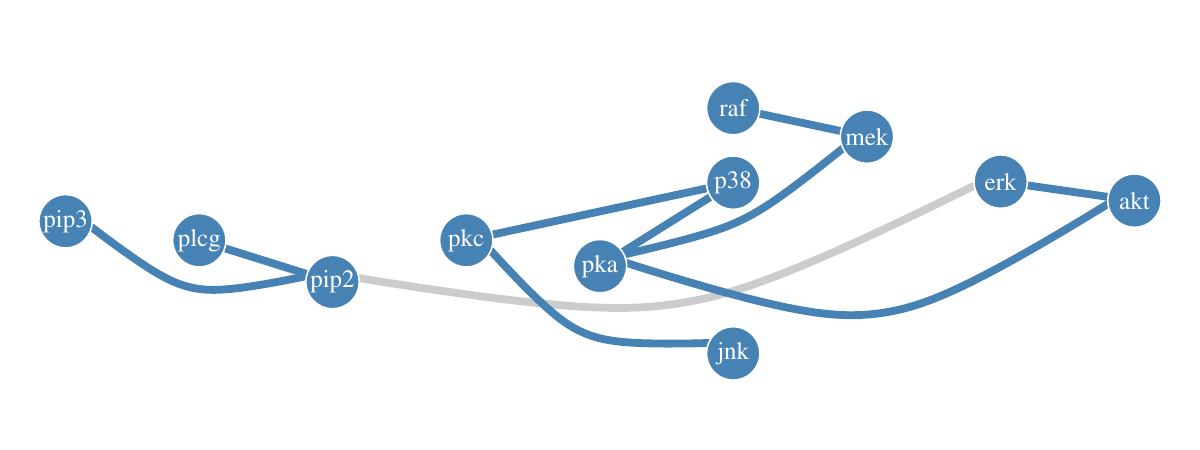}
 \includegraphics[width=0.5\linewidth]{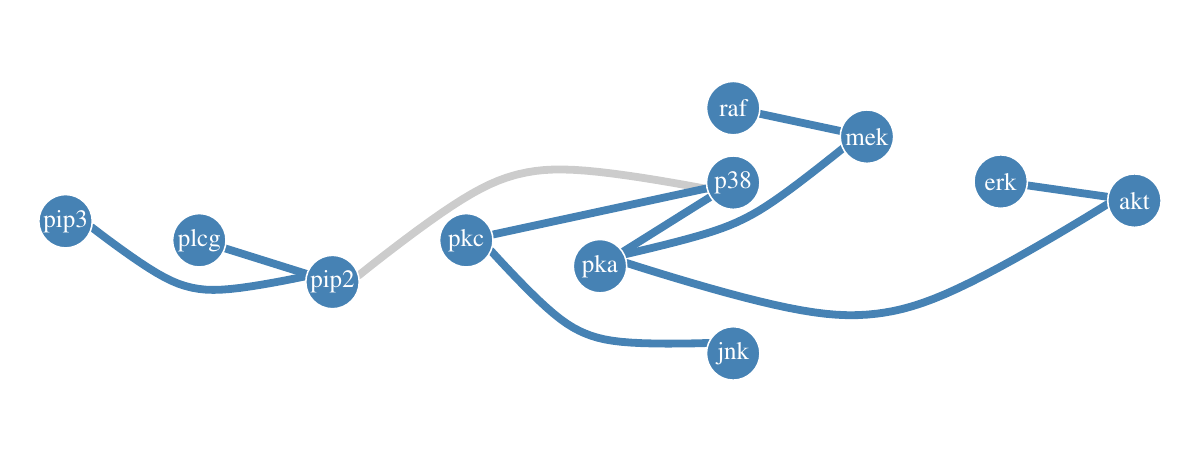}
 }
 
 \subfloat[Posterior probabilities for the edges in the tree model (with change of prior probability to $\lambda_{kl} = 1/2$ for all edges). 
 \label{RAF_posterior}]{\centering
 \includegraphics[width=0.75\linewidth]{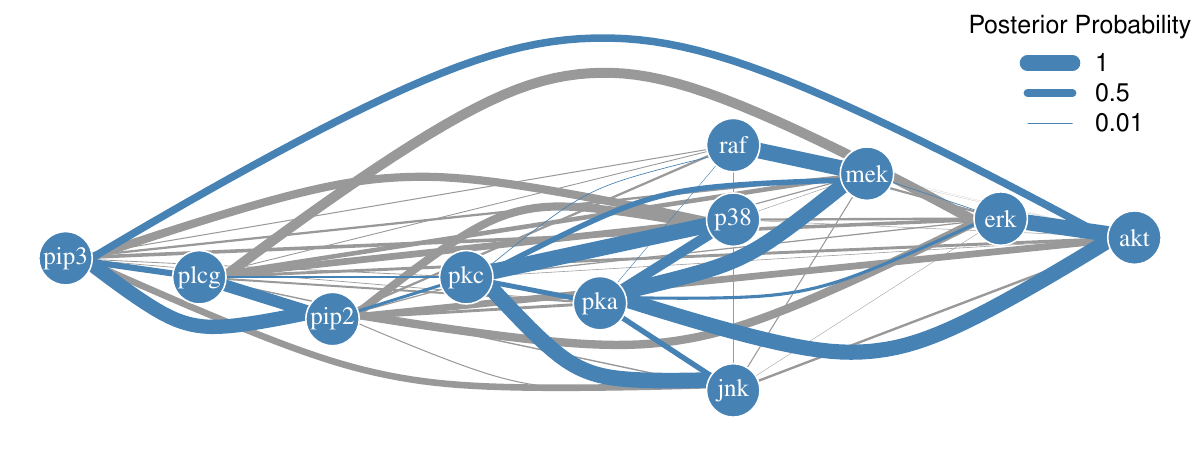}
 \includegraphics[width=0.25\linewidth]{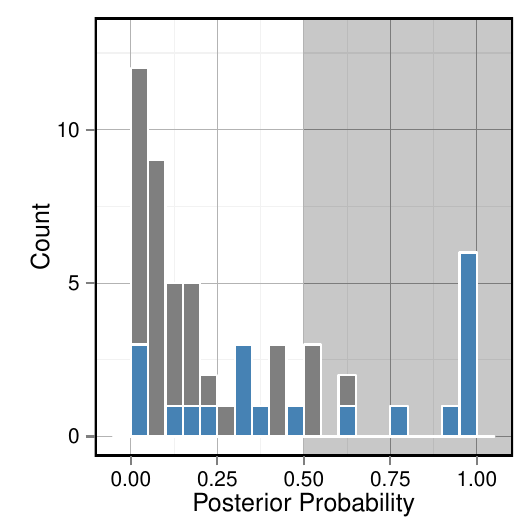}}
     \caption{Graphical representation of the results obtained on one of the five data sets. The edges of the golden standard network are colored in blue.}
     \label{res_RAF_particular}
 \end{figure}

\subsection{Data}
In flow cytometry experiments, cells are suspended in a stream of fluid and go through a laser beam one at a time. Different parameters are then measured on each cell by recovering the light that is reemitted by diffusion or fluorescence. We are interested in the activation levels (also called phosphorylation levels) of the involved proteins and phospholipids. Such experiments typically produce samples of several thousands observations. Since all biological network inference problems are not met by such a profusion of data, \cite{Werhli2006} sampled down 5 samples of size $n=100$ from the data provided by \cite{Sachs2005}. We discretised each sample into r=3 bins and performed the inference on each of them with our algorithm (Tree) and the MCMC sampling in DAGs algorithm (DAG), as described in the previous section. The accuracy of the inference was once again assessed by the area under the ROC and PR curves, averaged on all 5 samples.

\subsection{Results}


With the DAG approach, we got average areas under the ROC and PR curves of 0.767 and 0.725 respectively (with standard deviation of 0.068 and 0.070). With trees, we respectively got 0.729 and 0.690 for these areas (with standard deviation of 0.047 and 0.051). The DAG approach seems to perform better than our inference based on trees. These results qualify those of the previous section. Nonetheless, we would like to make the following points. While not being as accurate, our approach still provides good results and might in fact be more adapted to bigger problems where MCMC sampling can hardly be contemplated. Moreover, unlike the simulation study, the gold standard network against which the accuracy of the inference is assessed here, shown in Figure \ref{RAF_network}, is not perfectly known and may still differ quite considerably from the truth. 

Figure \ref{res_RAF_particular} gives a graphical representation of the results obtained on one of the five data sets, offering a more detailed overview. We note that the gold standard network as defined here has 20 edges. The two likeliest trees in the posterior tree distribution are given in Figure \ref{RAF_MAP}. Both trees have 9 true positives out of the $p - 1 = 10$ edges they respectively selected. As expected, most of these edges also have strong posterior probabilities (Figure \ref{RAF_posterior}). When the prior probabilities of all edges is brought back to $1/2$, we get $13$ edges with posterior probabilities strictly greater than $1/2$, among which the same true positives as in the MAP estimate. More generally, one could consider using the histogram of posterior probabilities to empirically find a more appropriate cut-off.

We did not represent the empirical edge frequencies obtained for DAGs since prior appearance probabilities could not be easily accounted for in this case, thus making direct comparison with posterior edge probabilities in trees impossible.

As a conclusion, these results lead us to believe that it might be preferable to favour inference using DAGs for small problems. When that is no longer possible in a reasonable amount of time, performing exact inference in a model based on trees is a computationally efficient alternative that can be used at a limited cost on the accuracy.

%% file: tex/appendix.tex
\section*{Appendix}

\setcounter{lemma}{0}

\LS{}{

\begin{proof}[Proof of Lemma \ref{lemma:Zd2}]

Let $\overline{Q}$ be the matrix obtained from $Q$ when row and column $u$ are removed. Notice that $\overline{Q} = \left[\Delta^{\{u\}}\right]^{-1}$. For convenience, we also let $R \defeq \Delta^{\{u\}} = \overline{Q}^{-1}$. The rows and columns of $\overline{Q}$ and $R$ are indexed by $\overline{V} \defeq V \setminus \{u\}$.

\noindent Let $k, l_1, l_2$ be pairwise distinct vertices in $V$. Using Theorem \ref{th:MT} and Lemma 3 of \citep{Meila06}, we get that
\begin{align*}
\frac{\partial^2 Z}{\partial\omega_{kl_1}\partial\omega_{kl_2}} &= \frac{\partial^2 |R|}{\partial\omega_{kl_1}\partial\omega_{kl_2}} \\
 & = \frac{\partial}{\partial\omega_{kl_1}}\left( |R|\cdot M_{kl_2} \right) \\
 & = |R|\cdot\left[M_{kl_1} M_{kl_2} + \frac{\partial M_{kl_2}}{\partial\omega_{kl_1}} \right]
\end{align*}
Assume that $u \not\in \{k,l_1,l_2\}$. Then $M_{kl_2} = \overline{Q}_{kk} + \overline{Q}_{l_2l_2} - 2\overline{Q}_{k,_2}$ and
\begin{align*}
\frac{\partial \overline{Q}_{kk}}{\partial\omega_{kl_1}} = \sum_{i,j \in \overline{V}} \frac{\partial \overline{Q}_{kk}}{\partial R_{ij}}\frac{\partial R_{ij}}{\partial\omega_{kl_1}} = - \sum_{i,j \in \overline{V}} \overline{Q}_{ki}\overline{Q}_{jk}\frac{\partial R_{ij}}{\partial\omega_{kl_1}} = - \left(\overline{Q}_{kk} - \overline{Q}_{kl_1}\right)^2
\end{align*}
where the last identity is obtained by noticing that the only terms of $R = \Delta^{\{u\}}$ that depend on $\omega_{kl_1}$ are $R_{kl_1}$, $R_{l_1k}$, $R_{kk}$ and $R_{l_1l_1}$. We similarly obtain that
\begin{align*}
\frac{\partial \overline{Q}_{l_2l_2}}{\partial\omega_{kl_1}} & = - \left(\overline{Q}_{l_1l_2} - \overline{Q}_{kl_2}\right)^2, \\
\frac{\partial \overline{Q}_{kl_2}}{\partial\omega_{kl_1}} & = \left(\overline{Q}_{kk} - \overline{Q}_{kl_1}\right)\left(\overline{Q}_{l_1l_2} - \overline{Q}_{kl_2}\right).
\end{align*}
Putting all pieces together, we get
\begin{align*}
\frac{\partial^2 Z}{\partial\omega_{kl_1}\partial\omega_{kl_2}}  & = |R|\cdot\left[M_{kl_1} M_{kl_2} - \left( \overline{Q}_{kk} - \overline{Q}_{kl_1} -  \overline{Q}_{kl_2} + \overline{Q}_{l_1l_2} \right)^2 \right], \\
& = Z\cdot M_{l_1l_2}^{(k)}.
\end{align*}
The cases $k = u$ and $l_2 = u$ are dealt with similarly.
\end{proof}}